\documentclass{article}

\usepackage{microtype}
\usepackage{graphicx}
\usepackage{subfigure}
\usepackage{booktabs} % for professional tables
\usepackage[table]{xcolor}
\usepackage{hyperref}
\usepackage{diagbox}
\usepackage{subcaption}
\usepackage{diagbox}
\usepackage{amsmath}
\usepackage{amssymb}
\usepackage{mathtools}
\usepackage{amsthm}
\usepackage{tabularx}
\usepackage{booktabs}  
\usepackage[capitalize,noabbrev]{cleveref}
\usepackage{bbm}          % \mathbbm{1} indicator function
\usepackage{subcaption}   % \subcaption in tables
\usepackage{diagbox}      % \diagbox for table headers
\usepackage{tabularx}     % tabularx environment
\usepackage{booktabs}     % \toprule \midrule \bottomrule
\usepackage{multirow}     % \multirow
\usepackage[table]{xcolor} 
\usepackage[accepted]{icml2025}
\usepackage{float}
\usepackage{amsmath}
\usepackage{amssymb}
\usepackage{mathtools}
\usepackage{amsthm}
\usepackage{tabularx}
\usepackage{booktabs}  
\usepackage[capitalize,noabbrev]{cleveref}
\usepackage{bbm}          % \mathbbm{1} indicator function
\usepackage{subcaption}   % \subcaption in tables
\usepackage{diagbox}      % \diagbox for table headers
\usepackage{tabularx}     % tabularx environment
\usepackage{booktabs}     % \toprule \midrule \bottomrule
\usepackage{multirow}     % \multirow
\usepackage[table]{xcolor} % \rowcolor
\theoremstyle{plain}

\theoremstyle{definition}

\theoremstyle{remark}

\usepackage[textsize=tiny]{todonotes}
\usepackage{multirow}
\icmltitlerunning{GEASS: Gated Evidence-Adaptive Selective Caption Trust for Vision-Language Models}
\makeatletter
\renewcommand\printAffiliationsAndNotice[1]{%
    \vskip 0.2in
    \noindent
    \begin{minipage}{\textwidth}
    \centering
    \@authorlistwithaffil
    \end{minipage}
    \vskip 0.2in
}
\newcommand\@authorlistwithaffil{%
    \def\and{\\}%
    \let\@affilsep\relax
    \@for\@auth:=\icml@authorlist\do{%
        \expandafter\@showauthor\@auth\\%
    }%
}
\def\@showauthor#1\\#2\\{%
    #1\\%
    \small\itshape\icml@getaffil{#2}\\%
    \ifx\icml@correspondingauthor\@empty\else
        \small\texttt{\icml@correspondingauthor}\\[2pt]
    \fi
}
\def\icml@getaffil#1{%
    \@ifundefined{icml@affil@#1}{}{%
        \csname icml@affil@#1\endcsname
    }%
}
\makeatother

\begin{document}

\twocolumn[
 \icmltitle{GEASS: Gated Evidence-Adaptive Selective Caption Trust for Vision-Language Models}

 % 横向作者列表（带单位上标）
 \vskip 0.1in
 \centering
 {\large
   Zeshang Li$^{1,2}$\quad
   Shuoyang Zhang$^{1}$\quad
 }\\
 \vskip 0.05in
 % 单位列表（与作者上标对应）
 \small
 $^{1}$University of International Relations    \\
 $^{2}$(Corresponding author email: \texttt{zeshang.li@uir.edu.cn})\\

 \vskip 0.2in
]




 % otherwise use the standard text.

\begin{abstract}
Vision-Language Models (VLMs) hallucinate objects that are not present, and a growing line of work tries to curb this by feeding the model its own generated caption as auxiliary evidence---assuming that a caption, once available, is something to consume. We show this fails: naively appending a caption can lower accuracy rather than raise it, dropping Qwen2.5-VL-3B$^\dagger$ on HallusionBench by nearly ten points. To understand why, we build \textbf{GD-Probe}, a diagnostic set that pairs a global and a detail question on the same image, so that any difference in caption effect is attributable to the question alone. Caption utility proves to be a \emph{per-query} property: the same caption helps global questions and harms detail ones, through a single mechanism---an embedded caption competes with the image for attention and pulls the model's evidence onto its own text---whose sign is set by whether the caption \emph{covers} the queried content. Crucially, this regime is readable from quantities the decoder already emits, with no attention access or grounding. We turn this into \textbf{GEASS} (Gated Evidence-Adaptive Selective Caption Trust), a training-free, logit-level module that decides per query how much of the caption to trust, gating it by the clean path's confidence, weighting it by the entropy reduction it induces, and raising the evidence bar when the two pathways disagree. Across four VLMs and two benchmarks (POPE and HallusionBench), GEASS improves over both vanilla inference and contrastive decoding under a single fixed setting, adding only two forward passes and no parameters.
\end{abstract}
\section{Introduction}
\label{sec:intro}

Vision-Language Models (VLMs)~\cite{wang2024qwen2, bai2025qwen25vl, liu2024improved, chen2024expanding} have become strong general-purpose systems for visual question answering, multimodal reasoning, and grounded generation. They nonetheless hallucinate: they assert objects, attributes, and relations that the image does not contain, and state these errors with the same fluency and confidence as their correct answers~\cite{li2023evaluating, augustin2025dash, guan2024hallusionbench}. A confidently wrong answer is hard to separate from a confidently right one, and that is what turns hallucination from a surface defect into a reliability problem.

\begin{figure}[t]
\centering
% Teaser. Replace path if needed (e.g. image/caption1.png).
\includegraphics[width=\linewidth]{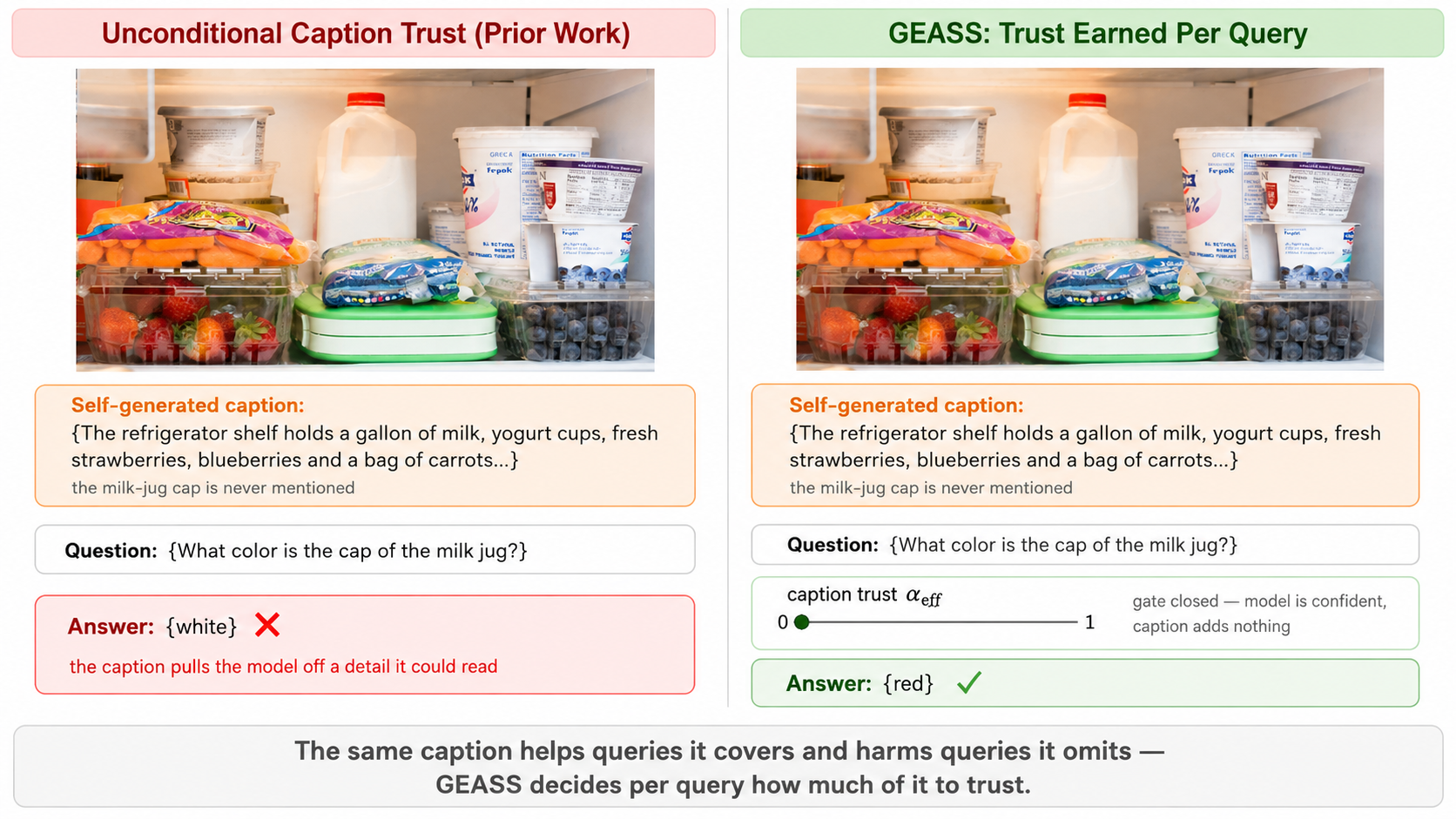}
\caption{Conventional caption-augmented inference (left) trusts the caption
unconditionally: anchored to a description that never mentions the milk-jug
cap, the model abandons a detail it could read and answers from the caption.
GEASS (right) treats trust as something earned per query: the clean path is
confident and the caption adds no information, so the gate stays closed and
the visual answer is preserved.}
\label{fig:teaser}
\end{figure}

A common explanation traces hallucination to the query-driven nature of visual reasoning: in answering a question, a VLM concentrates on the regions the question points to and overlooks evidence outside that scope~\cite{huang2024opera, tu2026attention}. One family of methods stays inside the model, either retraining it for better grounding~\cite{liu2023mitigating, sun2024aligning, yu2024rlhf} or intervening at decoding through contrastive decoding or attention manipulation~\cite{leng2024mitigating, huang2024opera, wang2025mitigating, tu2026attention}; such methods rebalance information the model already encoded but cannot supply evidence it never attended to. A second family adds that missing evidence from outside, in the form of an image caption: produced under a generic ``describe this image'' instruction, a caption gives a global account of the scene and can surface content that question-conditioned attention skips~\cite{hu2023promptcap, zhang2025scra, chen2024sharegpt4v, kapuriya2025enhancing}. Methods in this line generate a caption and feed it into reasoning as additional evidence, on the shared premise that a caption, once available, is something the model should consume.

This premise does not hold in practice. Appending a self-generated caption to the input lowers the accuracy of Qwen2.5-VL-3B~\cite{bai2025qwen25vl} on HallusionBench~\cite{guan2024hallusionbench} from 61.19 to 51.31 (Table~\ref{tab:caption_effect}), even though the caption only adds information. The drop is not uniform: the same caption that corrects one question drives the model to the wrong answer on another. A caption's usefulness is therefore a property of the \emph{caption--question pair}, not of the caption alone, and consuming captions unconditionally is the wrong default.

\begin{table}[t]
\caption{Effect of self-generated captions on HallusionBench (Qwen2.5-VL-3B$^\dagger$). Naively embedding a caption degrades accuracy by nearly ten points, despite supplying additional information.}
\label{tab:caption_effect}
\vspace{8pt}
\centering
\setlength{\tabcolsep}{8pt}
\begin{tabular}{lc}
\toprule
Caption Condition & HallusionBench \\
\midrule
No caption (Baseline) & 61.19 \\
Self-generated caption & 51.31 \\
\bottomrule
\end{tabular}
\end{table}

We take this as a reason to change the question. Rather than asking how to produce a better caption, which existing work pursues at the cost of stronger captioners or external verifiers, we ask:

\begin{quote}
\emph{Given a caption, when should a VLM trust it, and how much, on a particular query?}
\end{quote}

This leaves the caption fixed and regulates only how its content is used, which is cheaper than improving caption quality and orthogonal to it.

To answer it, we examine how an embedded caption changes the model's behavior and find a single mechanism. A caption does not sit beside the model's reasoning as an optional reference; it competes with the image for attention and draws the model's evidence onto its own text (Figure~\ref{fig:teaser}). Whether this helps is decided by what the caption holds for the query. Captions are reliable on the global layout of a scene but, produced by a saliency-weighted scan, omit small and peripheral objects. On global questions the caption supplies what query-driven attention missed and helps; on detail questions it pulls the model away from a region it could have read and replaces it with text that is silent about the detail, overturning an answer the model already had.

The same view shows how to act on a caption without inspecting attention or grounding, since this regime is visible in quantities a decoder already produces: the model's confidence in its caption-free answer, the entropy reduction the caption induces, and whether the caption overturns that answer. We turn these into \textbf{GEASS} (\textbf{G}ated \textbf{E}vidence-\textbf{A}daptive \textbf{S}elective Caption Tru\textbf{st}), a training-free procedure that runs the model with and without the caption and fuses the two logit distributions with a per-query weight from three signals: a \emph{confidence gate} that consults the caption only when the model is uncertain, an \emph{information-gain} weight that scales it by how much it sharpens the prediction, and a \emph{perceptual override guard} that, when the caption disagrees, demands evidence in proportion to how confident the model's visual answer was. The first two decide whether and how much to listen; the third makes confirming the model cheap and overturning it expensive.

GEASS adds no parameters and no architectural change, costs one caption-generation pass (amortizable across queries on the same
image) plus one extra answer-step forward pass per query, and applies to any VLM that exposes decoding logits. Across four VLMs and two benchmarks---POPE~\cite{li2023evaluating} and HallusionBench~\cite{guan2024hallusionbench}---it improves over both vanilla inference and a strong contrastive-decoding baseline under a single fixed hyperparameter setting.

Our contributions are summarized as follows:

\begin{itemize}
\item \textbf{Caption utility is per-query, and we explain why.} The same caption helps questions that require global understanding and harms questions about fine detail. We trace this to one mechanism: an embedded caption draws the model's attention from the image onto its own text, which helps when the caption covers the query and harms when it omits the queried detail.

\item \textbf{GEASS, a training-free trust-regulation method.} A logit-level procedure that sets per-query caption trust from three decoding signals---a confidence gate, an information-gain weight, and a perceptual override guard---improving over vanilla inference and contrastive decoding across four VLMs and two benchmarks (POPE, HallusionBench), at about $2.3\times$ the cost of greedy decoding and with no per-model tuning.
\end{itemize}
\section{Related Work}
\label{sec:related}

\subsection{Inference-Time Hallucination Mitigation}
Object hallucination is a well-documented reliability problem in VLMs~\cite{liu2024improved, dai2023instructblip, bai2025qwen25vl, chen2024expanding}, measured by object-existence polling~\cite{li2023evaluating}, hallucinated-object rates in captioning~\cite{rohrbach2018object}, and broader perception stress tests~\cite{augustin2025dash, guan2024hallusionbench}. Mitigation splits along training and inference. Training-time methods improve grounding with curated data or alignment objectives~\cite{liu2023mitigating, sun2024aligning, yu2024rlhf}, at substantial compute and supervision cost. Inference-time methods avoid retraining and act during generation, through external verification~\cite{yin2024woodpecker}, attention manipulation~\cite{tu2026attention, tong2024eyes}, or contrastive decoding~\cite{leng2024mitigating, huang2024opera, wang2025mitigating, park2025second, sarkar2025mitigating}.

The line closest to ours operates at the logit level: it builds a degraded counterpart of the input and subtracts its influence from the output. VCD~\cite{leng2024mitigating} contrasts against a noise-corrupted image, and CODE~\cite{kim2024code} contrasts against the model's own description. What unites them is the assumption that the contrasted signal is uniformly misleading and should be removed. A self-generated caption does not fit this picture: it is mixed-quality---reliable on global content and weak on detail---so subtracting it also discards the evidence it does carry. We therefore regulate a caption's influence per query rather than contrast it away.

\subsection{Captions as Auxiliary Evidence in VLMs}
A complementary line supplies the missing visual evidence as text and reasons over it. Recent training-free methods generate a description and feed it back into the model: GeReA~\cite{ma2024gerea} prompts a multimodal model for question-aware captions and reasons over them for knowledge-based VQA, CCoT~\cite{mitra2024compositional} generates a scene-graph description as an intermediate reasoning step, and VDGD~\cite{ghosh2025visual} prepends a detailed description to the instruction and grounds decoding toward it. Other methods feed captions to VLMs through task-aware captioning~\cite{hu2022promptcap}, language-model feedback~\cite{liu2024zvqaf}, higher-quality caption data~\cite{chen2024sharegpt4v, zhang2025scra, kapuriya2025enhancing}, or caption-conditioned attention~\cite{li2026cast, li2025cai}.

These methods differ in mechanism but share a premise: once a description is available it is beneficial and should be consumed, so progress means producing better descriptions or grounding the model more tightly to them. Even work that notes a caption can introduce irrelevant content frames the remedy as a better caption, not as variable trust. We question the premise itself. As Section~\ref{sec:analysis} shows, an embedded caption helps or harms depending on the question it is paired with; the open problem is therefore not how to generate a better caption, but when, and how much, an existing one should be trusted.
\section{Understanding Caption Utility in VLM Inference}
\label{sec:analysis}

We study how an embedded caption reshapes VLM behavior in three steps: when it helps or hurts (\S\ref{sec:phenomenon}), the mechanism behind both outcomes (\S\ref{sec:mechanism}--\S\ref{sec:coverage}), and whether that mechanism is readable at inference (\S\ref{sec:signals}).

\paragraph{Setup.}
We use four VLMs: InternVL2-8B~\cite{chen2024expanding}, InternVL3-8B~\cite{zhu2025internvl3}, Qwen2.5-VL-3B~\cite{bai2025qwen25vl}, and a reasoning-augmented variant Qwen2.5-VL-3B$^\dagger$, fine-tuned with chain-of-thought supervision so that it emits explicit \texttt{<think>} reasoning before its answer. Each caption is generated by the same model from ``Describe this image in detail'' (greedy, fixed per image) and embedded alongside the image. We measure the aggregate caption effect on HallusionBench~\cite{guan2024hallusionbench}, and base our controlled analyses on COCO~\cite{lin2014coco}, where object-level grounding is available.

\subsection{Caption Utility Is Query-Dependent}
\label{sec:phenomenon}

Embedding a self-generated caption lowers HallusionBench accuracy by nearly ten points (Table~\ref{tab:caption_effect}). The aggregate is misleading: per query, the same caption fixes some questions and breaks others, so the drop is the net of two opposing effects that depend on what each question asks.

\paragraph{Global vs.\ detail questions.}
We split questions by the visual evidence they require: \emph{global} questions depend on scene-level content (layout, the dominant objects, their relations, or a count), while \emph{detail} questions hinge on a single small or peripheral object. To study the split under controlled conditions, we construct \textbf{GD-Probe}, a diagnostic set of COCO images, each paired with two questions---one global and one detail---where the detail question targets an object whose ground-truth box covers under $5\%$ of the image. Pairing the two question types on the same image holds the image and its caption fixed, so any difference in caption effect is attributable to the question rather than the picture. Because GD-Probe is built on COCO, its object boxes also support the attention and coverage analyses in \S\ref{sec:mechanism}--\S\ref{sec:coverage}; we report its scale and give example questions in Appendix~\ref{app:gdprobe}.

\begin{figure}[t]
\centering
\includegraphics[width=0.95\linewidth]{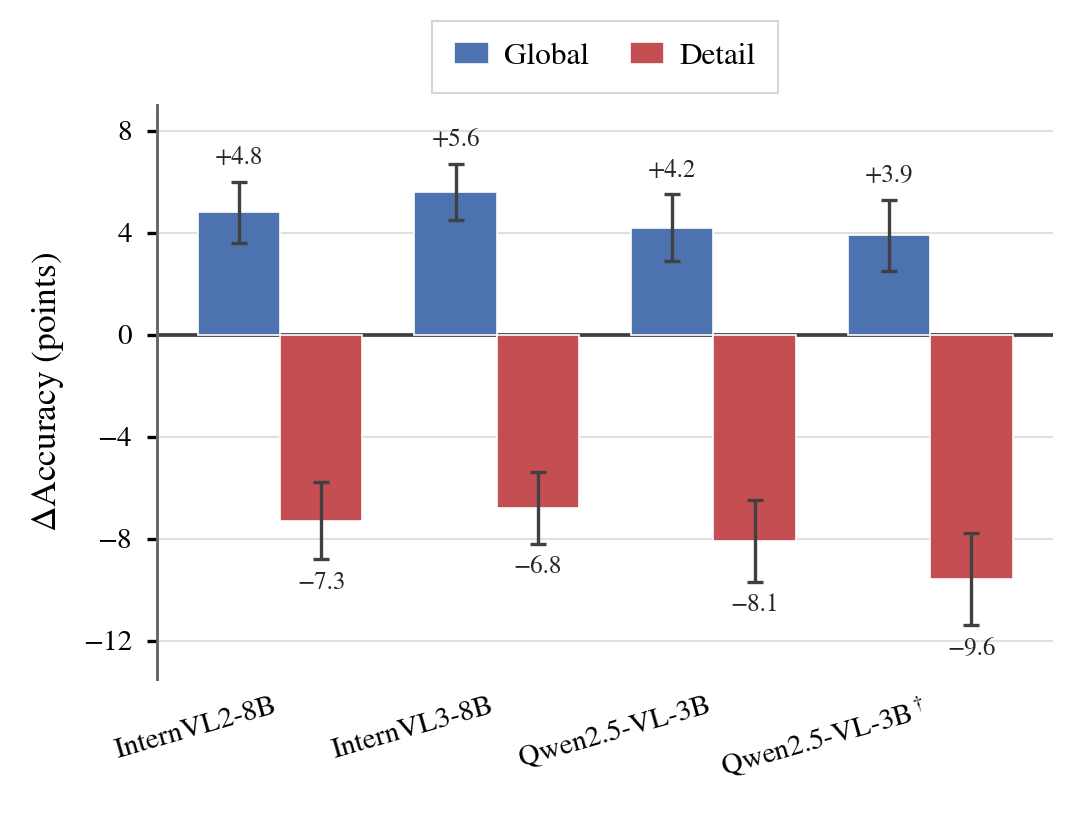}
\caption{Accuracy change from caption injection on GD-Probe, split by question type, across four VLMs. Captions consistently help global questions and hurt detail questions; whether a benchmark's aggregate rises or falls depends on its mix of the two (cf.\ Table~\ref{tab:caption_effect}).}
\label{fig:global_detail}
\end{figure}

\paragraph{Captions help global and hurt detail.}
On GD-Probe, across all four models, caption injection raises accuracy on global questions and lowers it on detail questions (Figure~\ref{fig:global_detail}). The aggregate drop in Table~\ref{tab:caption_effect} is then simply the net of the two: on this benchmark the harm on detail questions outweighs the help on global ones, so accuracy falls overall even though some questions improve. A caption's value is therefore set by the question it is paired with, not by the caption alone.

Two controls confirm this is the real effect. First, the harm removes answers the model already had: restricting to detail questions each model answers correctly \emph{without} a caption, embedding the caption flips a substantial fraction back to wrong (detail erosion rate, Table~\ref{tab:detail_erosion}). Second, the effect is content-driven, not a length artifact: against three length-matched replacements---neutral filler, an unrelated image's caption, and the word-shuffled caption---only the genuine caption improves global accuracy, and only the genuine caption degrades detail beyond the neutral baseline (Table~\ref{tab:control}). Caption utility is thus a property of the \emph{caption--question pair}: the same caption supplies what a global question needs while overturning what a detail question already had. The rest of this section asks by what mechanism.

\begin{table}[t]
\centering
\caption{Detail erosion. On the subset of detail questions each model answers correctly \emph{without} a caption ($N_{\text{correct}}$), embedding the caption flips a substantial fraction to incorrect. DER $=$ detail erosion rate (lower is better).}
\label{tab:detail_erosion}
\vspace{8pt}
\footnotesize
\setlength{\tabcolsep}{4pt}
\begin{tabular}{lccc}
\toprule
Model & $N_{\text{correct}}$ & Acc.\ w/ caption (\%) & DER (\%) \\
\midrule
InternVL2-8B            & 312 & 84.0 & 16.0 \\
InternVL3-8B            & 298 & 85.6 & 14.4 \\
Qwen2.5-VL-3B           & 305 & 82.3 & 17.7 \\
Qwen2.5-VL-3B$^\dagger$ & 289 & 79.1 & 20.9 \\
\bottomrule
\end{tabular}
\end{table}

\begin{table}[t]
\centering
\caption{Content-vs-length controls on Qwen2.5-VL-3B$^\dagger$. All text conditions are length-matched. Only the genuine self-generated caption improves global accuracy, while it degrades detail accuracy beyond the neutral-text baseline, isolating a content-driven effect from a sequence-length effect.}
\label{tab:control}
\vspace{8pt}
\small
\begin{tabular}{lccc}
\toprule
Input condition & Len.\ (tok.) & Global (\%) & Detail (\%) \\
\midrule
No caption          & 0  & 63.5 & 68.0 \\
Neutral filler      & 56 & 63.2 & 66.4 \\
Unrelated caption   & 56 & 62.8 & 65.3 \\
Shuffled caption    & 56 & 63.0 & 66.0 \\
Self caption        & 56 & \textbf{67.4} & \textbf{58.4} \\
\bottomrule
\end{tabular}
\end{table}

\subsection{Captions Displace Visual Evidence with Text}
\label{sec:mechanism}

We test one mechanism for the sign flip: an embedded caption competes with the image for attention, so the answer rests less on the image and more on the caption. We probe where the model attends, what its output copies, and what an intervention does.

\paragraph{Attention measurements.}
For the answer token we aggregate attention over heads in the upper layers (robustness in the appendix) and partition the context into image tokens $\mathcal{I}$, caption tokens $\mathcal{C}$, and the rest, with mass $a_j$ on token $j$. We define the Image and Caption Attention Shares,
\begin{equation}
\mathrm{IAS}=\frac{\sum_{j\in\mathcal{I}} a_j}{\sum_j a_j},
\qquad
\mathrm{CAS}=\frac{\sum_{j\in\mathcal{C}} a_j}{\sum_j a_j},
\end{equation}
and, using the queried object's box, the Target-Region Attention over the image patches $\mathcal{B}\subseteq\mathcal{I}$ that overlap the box,
\begin{equation}
\mathrm{TRA}=\frac{\sum_{j\in\mathcal{B}} a_j}{\sum_{j\in\mathcal{I}} a_j}.
\end{equation}
$\mathrm{IAS}$ and $\mathrm{CAS}$ say where the model looks at the image-vs-text level; $\mathrm{TRA}$ says whether it still looks at the region a detail question needs.

\begin{figure}[t]
\centering
\includegraphics[width=0.95\linewidth]{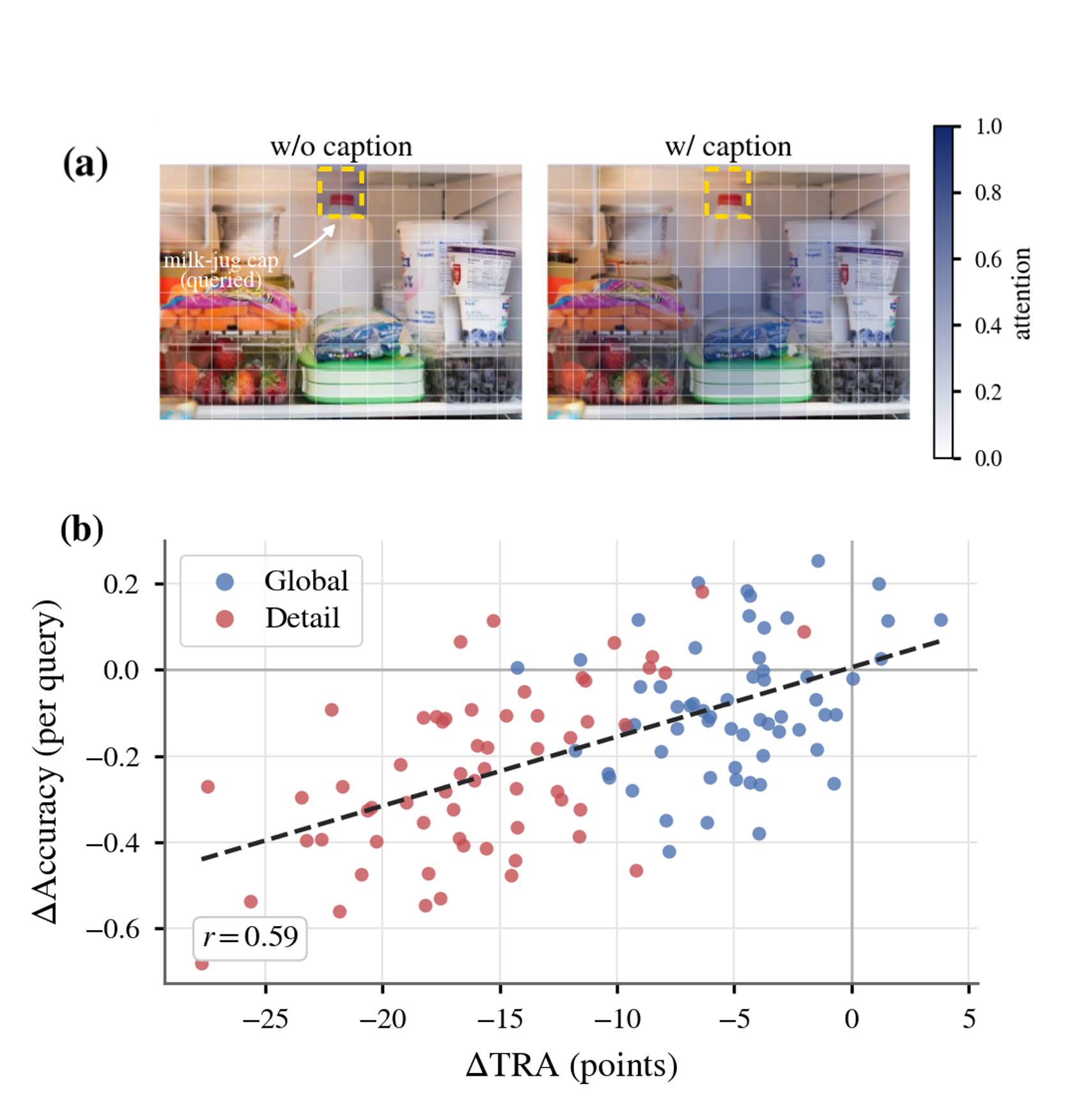}
\caption{\textbf{(a)} Answer-token attention as a grid overlay on a representative example (darker cell $=$ more attention). The detail question asks the colour of the milk-jug cap, an attribute the caption omits: without the caption (left) attention concentrates on the cap; with the caption (right) the cap lightens while attention spreads across the caption-mentioned scene. Illustrative single example; the quantitative claim is carried by Table~\ref{tab:attention_shift} and panel (b). \textbf{(b)} Across queries, the attention lost on the target region ($\Delta\mathrm{TRA}$) tracks the accuracy drop, most strongly on detail questions.}
\label{fig:attention}
\end{figure}

\paragraph{The caption pulls attention off the image.}
Adding the caption lowers $\mathrm{IAS}$ and diverts mass to the caption ($\mathrm{CAS}$) on both architectures (Table~\ref{tab:attention_shift}), and $\mathrm{TRA}$ on the queried object drops---far more for detail than global questions. The model stops looking exactly where detail questions hinge: the target region cools once the caption is added, and across queries the $\mathrm{TRA}$ drop tracks the accuracy drop (Figure~\ref{fig:attention}).

\begin{table}[t]
\centering
\caption{Attention reallocation under caption injection (answer-token attention, upper layers). Adding the caption lowers Image Attention Share ($\mathrm{IAS}$) and diverts mass to the caption ($\mathrm{CAS}$); Target-Region Attention ($\mathrm{TRA}$) on the queried object drops, and more so for detail questions ($\Delta\mathrm{TRA}=\mathrm{TRA}_{\text{cap}}-\mathrm{TRA}_{\text{clean}}$).}
\label{tab:attention_shift}
\vspace{8pt}
\scriptsize
\setlength{\tabcolsep}{6pt}
\begin{tabular}{lccccc}
\toprule
\multirow{2}{*}{Model} & $\mathrm{IAS}_{\text{clean}}$ & $\mathrm{IAS}_{\text{cap}}$ & $\mathrm{CAS}_{\text{cap}}$ & $\Delta\mathrm{TRA}$ & $\Delta\mathrm{TRA}$ \\
 & (\%) & (\%) & (\%) & global & detail \\
\midrule
InternVL2-8B   & 62.4 & 47.1 & 22.3 & $-3.8$ & $-14.6$ \\
Qwen2.5-VL-3B  & 58.0 & 43.5 & 26.1 & $-4.5$ & $-16.9$ \\
\bottomrule
\end{tabular}
\end{table}

\begin{figure}[t]
\centering
\includegraphics[width=0.95\linewidth]{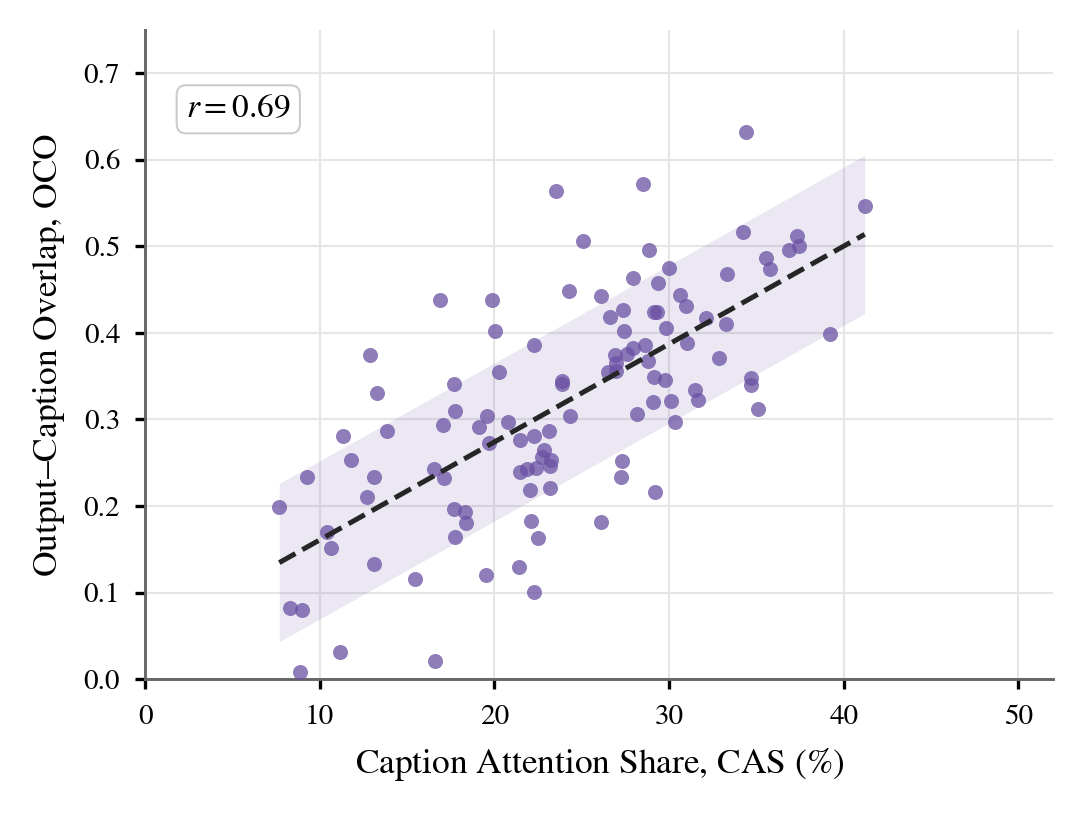}
\caption{Per-query Output--Caption Overlap ($\mathrm{OCO}$) against Caption Attention Share ($\mathrm{CAS}$). The two are positively correlated: outputs reproduce the caption in proportion to how much attention the query places on it, identifying lexical anchoring as the behavioral side of the attention shift.}
\label{fig:oco_cas}
\end{figure}

\paragraph{The shift surfaces in the output.}
If attention moves to the caption, the output should follow. The Output--Caption Overlap ($\mathrm{OCO}$, n-gram overlap between response and caption) rises with the caption and correlates with $\mathrm{CAS}$ across queries (Figure~\ref{fig:oco_cas}): lexical anchoring to the caption is the behavioral footprint of the attention shift, not a separate phenomenon.

\begin{figure}[t]
\centering
\includegraphics[width=0.95\linewidth]{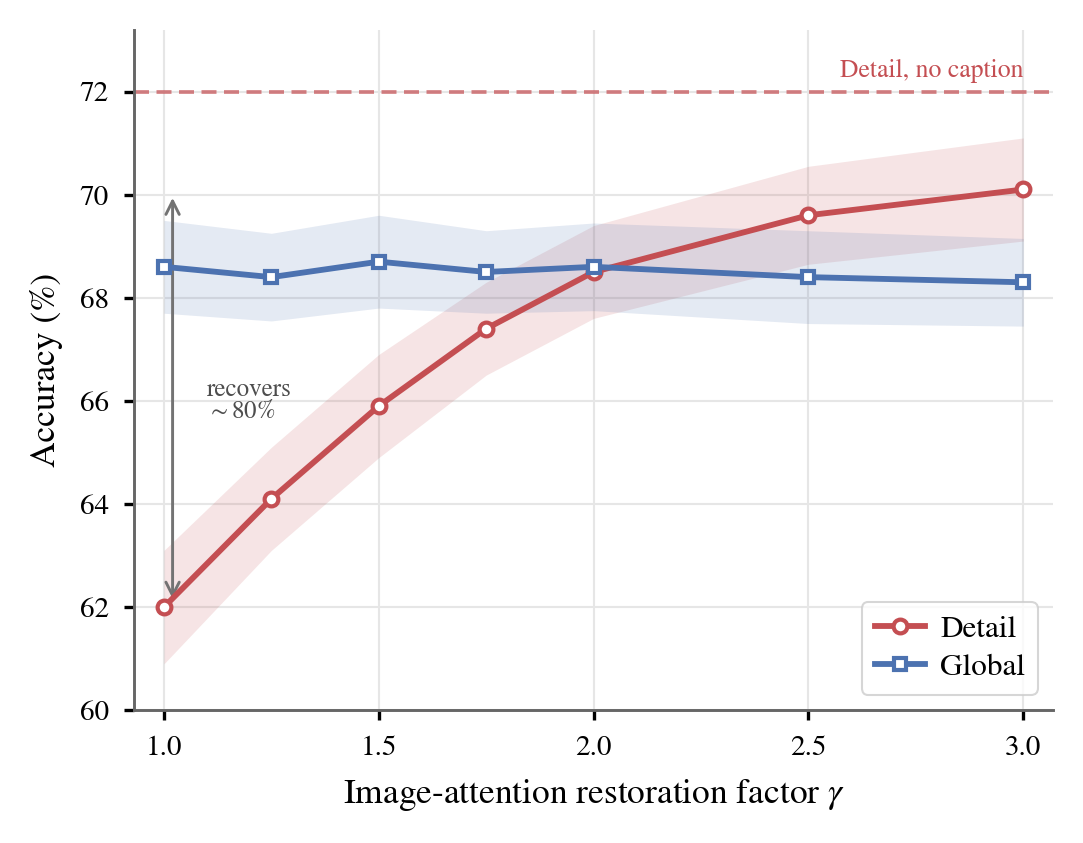}
\caption{Causal intervention. Rescaling answer-token attention back toward the image by a factor $\gamma$ ($\gamma{=}1$: unmodified caption condition) recovers detail accuracy monotonically while leaving global accuracy almost unchanged, establishing the attention shift as the cause of detail errors.}
\label{fig:gamma_sweep}
\end{figure}

\paragraph{The shift is causal.}
Maps are correlational, so we intervene: at the answer step we rescale attention toward the image by a factor $\gamma$ (down-weighting caption tokens) and renormalize, leaving the rest unchanged. As image attention is restored, detail accuracy rises monotonically and recovers much of the loss in Table~\ref{tab:detail_erosion}, while global accuracy is barely affected (Figure~\ref{fig:gamma_sweep}). The shift thus \emph{causes} the detail errors, and the caption's influence is a controllable lever---the lever GEASS pulls at the logit level, without per-step attention surgery. The redistribution is not harmful in itself; its effect depends on what the model now reads in the caption, which we examine next.

\subsection{Coverage Determines Whether the Shift Helps or Harms}
\label{sec:coverage}

What the shift costs or buys depends on \emph{coverage}: whether the caption contains the evidence the query needs. The global/detail split follows from how coverage is distributed.

\begin{figure*}[t]
\centering
\includegraphics[width=0.9\textwidth]{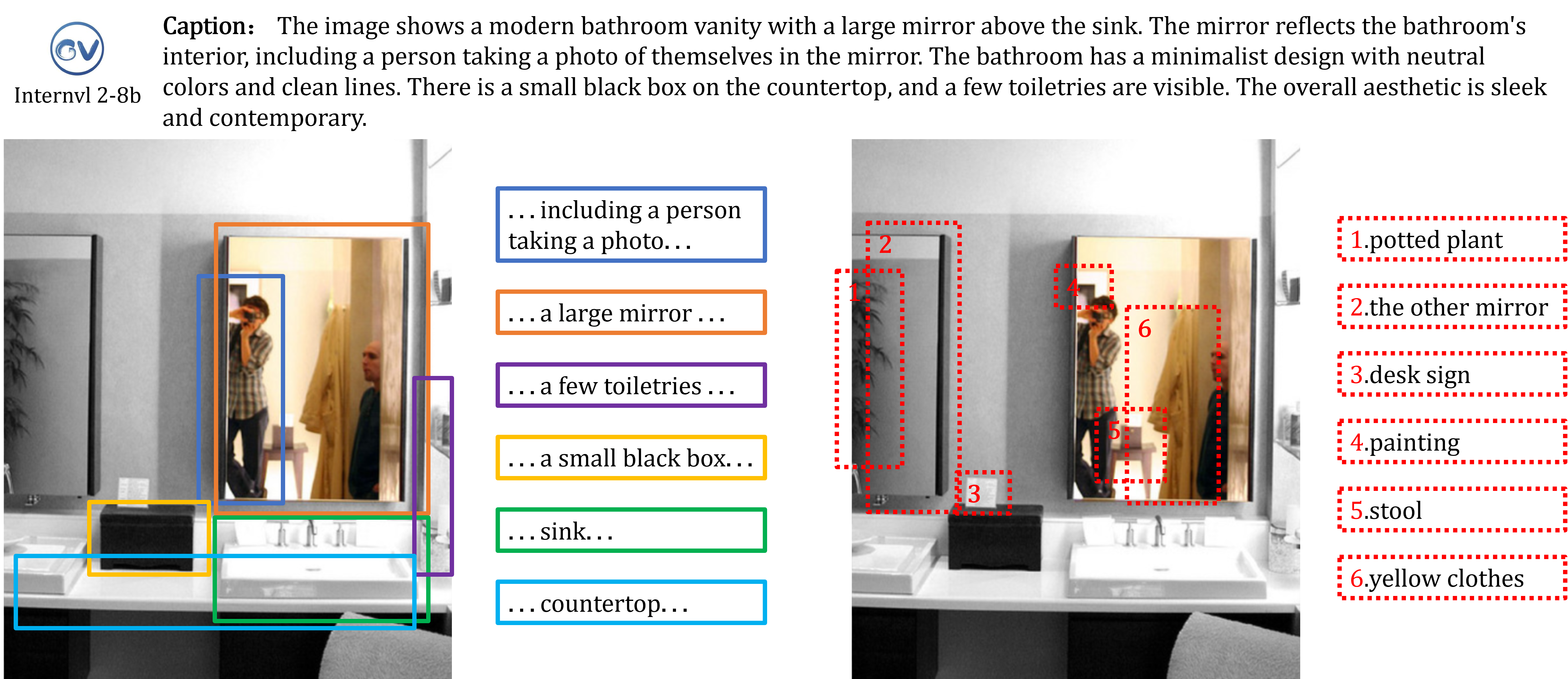}
\caption{A self-generated caption (top) names the salient foreground content of the scene---the large mirror, the sink and countertop, the person taking a photo, the toiletries, and the small black box (left, solid boxes)---while silently omitting peripheral or secondary objects: a potted plant, the second mirror, a desk sign, a wall painting, a stool, and the person's yellow clothing (right, dashed boxes). A detail question about any omitted object concerns content the caption never describes.}
\label{fig:coverage}
\end{figure*}

\begin{figure}[t]
\centering
\includegraphics[width=0.95\linewidth]{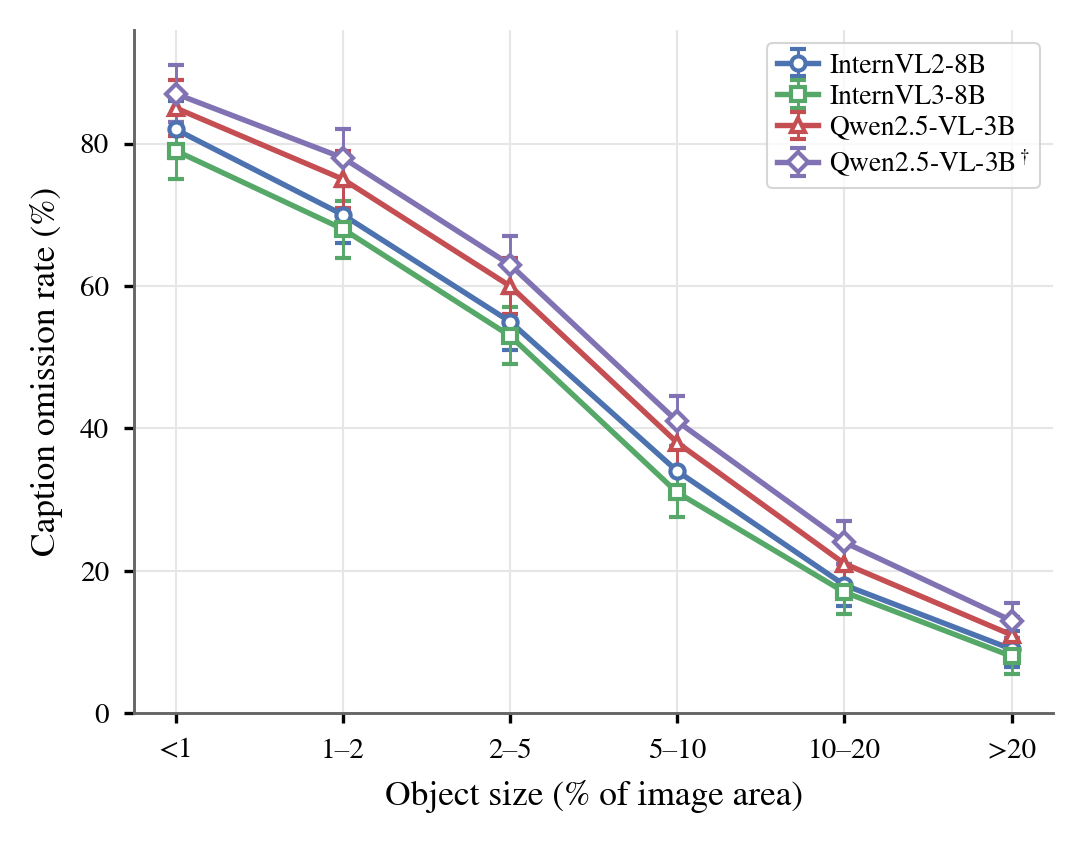}
\caption{Caption omission rate as a function of object size on COCO across four VLMs: smaller objects are dropped far more often, so detail questions usually concern content the caption never describes.}
\label{fig:omission}
\end{figure}

\paragraph{Captions cover the scene but omit the detail.}
Call a caption \emph{covering} for a query when the queried content $o_q$ is mentioned in it (matched with synonym expansion). A generic description prompt elicits a saliency-weighted scan that names the dominant, foreground content and passes over what is peripheral, secondary, or small (Figure~\ref{fig:coverage}). One clean, measurable axis of this bias is object size: across COCO, the omission rate grows monotonically as the queried object shrinks (Figure~\ref{fig:omission}). Detail questions therefore land mostly on omitted content, global questions on covered content.

\paragraph{The sign tracks coverage, not question type.}
Crossing question type with coverage (Table~\ref{tab:coverage_split}), the accuracy change follows coverage across all models: covered queries gain and omitted queries lose, whether global or detail. The detail harm is thus an omission effect, unavoidable for detail questions only because their targets are what captions routinely drop. This closes the account: a caption pulls attention onto its text (\S\ref{sec:mechanism}); when it covers the query the shift imports missed evidence and helps, and when it omits the query---the common case for detail---it trades a high-information region for silent text and overturns a correct answer. Even an accurate caption harms a query it does not cover, so the issue is trust, not quality. The mechanism, however, is diagnosed with grounding and attention a deployed system cannot use; the next section recovers the same regime from decoding alone.

\begin{table}[t]
\centering
\caption{Accuracy change from caption injection, crossed by question type and
caption coverage (covering vs.\ omitting the queried content). The \emph{sign}
of $\Delta$Acc is set by coverage, not question type---covered queries gain and
omitted queries lose for both global and detail---while the omitted-query loss
is consistently larger for detail, whose answer hinges on a specific region the
caption-induced attention shift abandons (\S\ref{sec:mechanism}), and is largest
for the reasoning-augmented Qwen2.5-VL-3B$^\dagger$. Per-cell sample counts are
in Table~\ref{tab:coverage_counts}.}
\label{tab:coverage_split}
\vspace{8pt}
\scriptsize
\setlength{\tabcolsep}{6pt}
\begin{tabular}{ll|cccc}
\toprule
\multirow{2}{*}{Type} & \multirow{2}{*}{Coverage} & \multicolumn{4}{c}{$\Delta$Acc (\%)} \\
 & & InternVL2 & InternVL3 & Qwen2.5 & Qwen2.5$^\dagger$ \\
\midrule
\multirow{2}{*}{Global} & Covered & $+6.8$ & $+7.4$ & $+6.1$ & $+5.9$ \\
                        & Omitted & $-5.4$ & $-4.8$ & $-6.2$ & $-6.9$ \\
\midrule
\multirow{2}{*}{Detail} & Covered & $+6.2$ & $+6.9$ & $+5.7$ & $+5.4$ \\
                        & Omitted & $-11.0$ & $-10.0$ & $-12.0$ & $-14.5$ \\
\bottomrule
\end{tabular}
\end{table}

\subsection{The Help/Harm Regime Is Readable at Decoding}
\label{sec:signals}

The analysis so far relied on measurements a deployed system cannot make at inference. Target-Region Attention needs the ground-truth box of the queried object; the causal test needs to read and rescale the model's internal attention and to pick its strength $\gamma$ per query; and many efficient attention kernels used in practice do not even expose the weights such edits require. These are diagnostic tools, not a method. We now show that the same help/harm regime can be read from quantities the decoder already outputs---no boxes, no attention internals, and no extra model.

\paragraph{Three signals from two passes.}
Running the model without and with the caption gives the clean and caption-conditioned distributions $p_{\text{clean}},p_{\text{cap}}$ at the answer step. From them we read the clean-path confidence
\begin{equation}
c=\max_{v} p_{\text{clean}}(v),
\end{equation}
the relative entropy reduction
\begin{equation}
r=\frac{H(p_{\text{clean}})-H(p_{\text{cap}})}{H(p_{\text{clean}})+\epsilon},
\end{equation}
and whether the answer changes, $\mathbb{1}[\arg\max p_{\text{clean}}\neq\arg\max p_{\text{cap}}]$. $c$ is low when the model missed what a global question needs and high when it resolved a detail; $r$ is large when the caption sharpens the prediction and near zero when it adds nothing. None needs a box or attention internals.

\begin{figure}[t]
\centering
\includegraphics[width=0.95\linewidth]{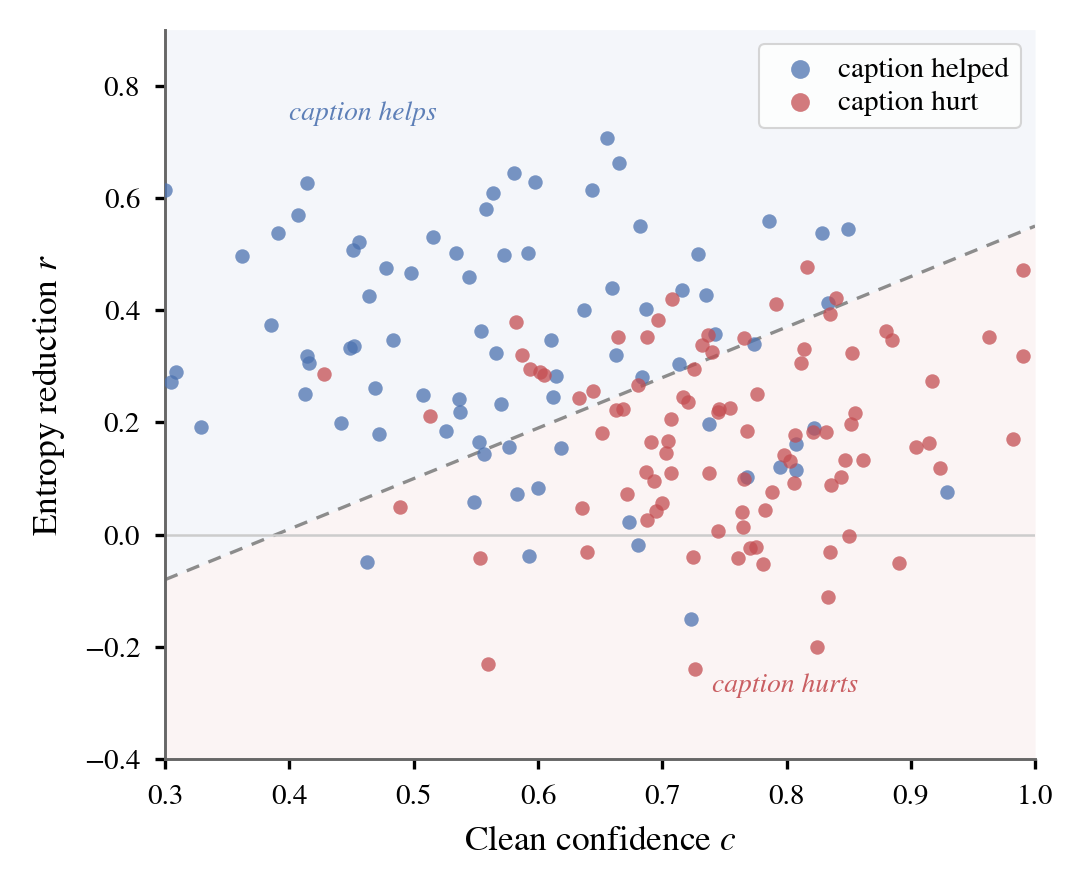}
\caption{Every query by clean confidence $c$ and entropy reduction $r$, colored by whether the caption helped or hurt. Beneficial cases concentrate at low $c$ and high $r$; harmful cases at high $c$ and low $r$. The regime diagnosed with attention and grounding is recovered from two decoding signals.}
\label{fig:signal_scatter}
\end{figure}

\paragraph{The cheap signals recover the mechanism.}
They track the directly measured mechanism (Table~\ref{tab:proxy}): $r$ is large on covered queries and near zero on omitted ones, recovering coverage without parsing the caption, and $c$ is low on global and high on detail questions. Plotting each query by $(c,r)$ and coloring by outcome (Figure~\ref{fig:signal_scatter}) separates the regimes---beneficial cases at low $c$/high $r$, harmful cases at high $c$/low $r$.

\begin{table}[t]
\centering
\caption{Decoding signals recover the mechanism. Mean entropy reduction $\bar{r}$ on caption-covered versus omitted queries (Section~\ref{sec:coverage} labels), and mean clean confidence $\bar{c}$ on global versus detail questions. $r$ tracks coverage and $c$ tracks where the model needs help, with no grounding or attention access.}
\label{tab:proxy}
\vspace{8pt}
\footnotesize
\setlength{\tabcolsep}{4pt}
\begin{tabular}{ll|cccc}
\toprule
\multicolumn{2}{l|}{Signal / split} & InternVL2 & InternVL3 & Qwen2.5 & Qwen2.5$^\dagger$ \\
\midrule
\multirow{2}{*}{$\bar{r}$} & Covered & $0.41$ & $0.43$ & $0.38$ & $0.36$ \\
                           & Omitted & $0.04$ & $0.03$ & $0.02$ & $-0.01$ \\
\midrule
\multirow{2}{*}{$\bar{c}$} & Global  & $0.57$ & $0.55$ & $0.60$ & $0.58$ \\
                           & Detail  & $0.80$ & $0.79$ & $0.83$ & $0.85$ \\
\bottomrule
\end{tabular}
\end{table}

\begin{figure}[t]
\centering
\includegraphics[width=0.95\linewidth]{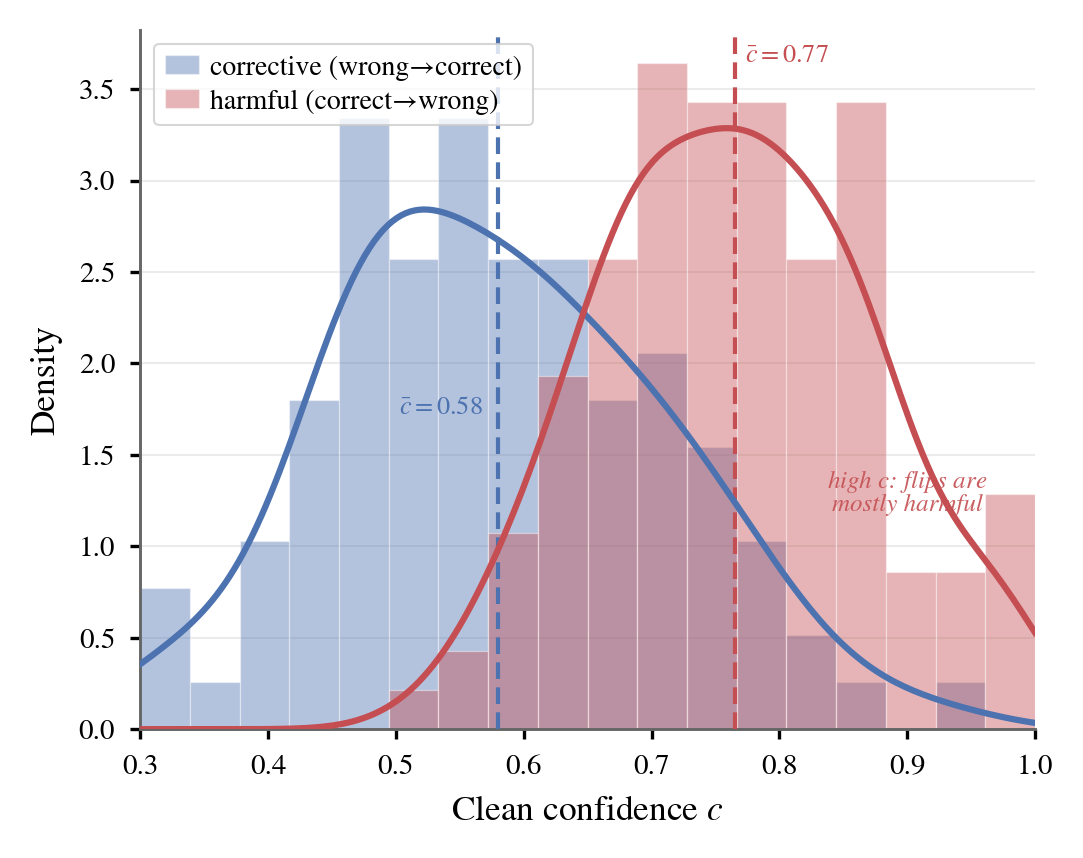}
\caption{Caption-induced answer flips by clean confidence $c$. Harmful flips (the caption overturns a correct visual answer) concentrate at high confidence, while corrective flips occur at low confidence, motivating an evidence bar for overrides that scales with $c$.}
\label{fig:flip_vs_conf}
\end{figure}

\paragraph{Overturning a confident answer is the risky move.}
A caption that changes the answer either corrects a wrong model or overturns a right one. Harmful flips concentrate where $c$ is high; corrective flips occur where the model was unsure (Figure~\ref{fig:flip_vs_conf}). The model's confidence in its own answer is the risk signal: the surer it was, the more an overturn is likely a distraction, and the more evidence a caption should have to supply to win.

\paragraph{Design principles.}
Each signal yields one principle: consult the caption only when $c$ is low (a confident model is resolving a detail the caption can only disturb); weight it by the coverage that $r$ reports; and require evidence for an overturn that grows with $c$, so confirming the model is cheap and overturning it expensive. These become the three components of GEASS---a confidence gate, an information-gain weight, and a perceptual override guard---formalized in Section~\ref{sec:method}.
\section{Method}
\label{sec:method}

\begin{figure*}[t]
\centering
\includegraphics[width=0.9\linewidth]{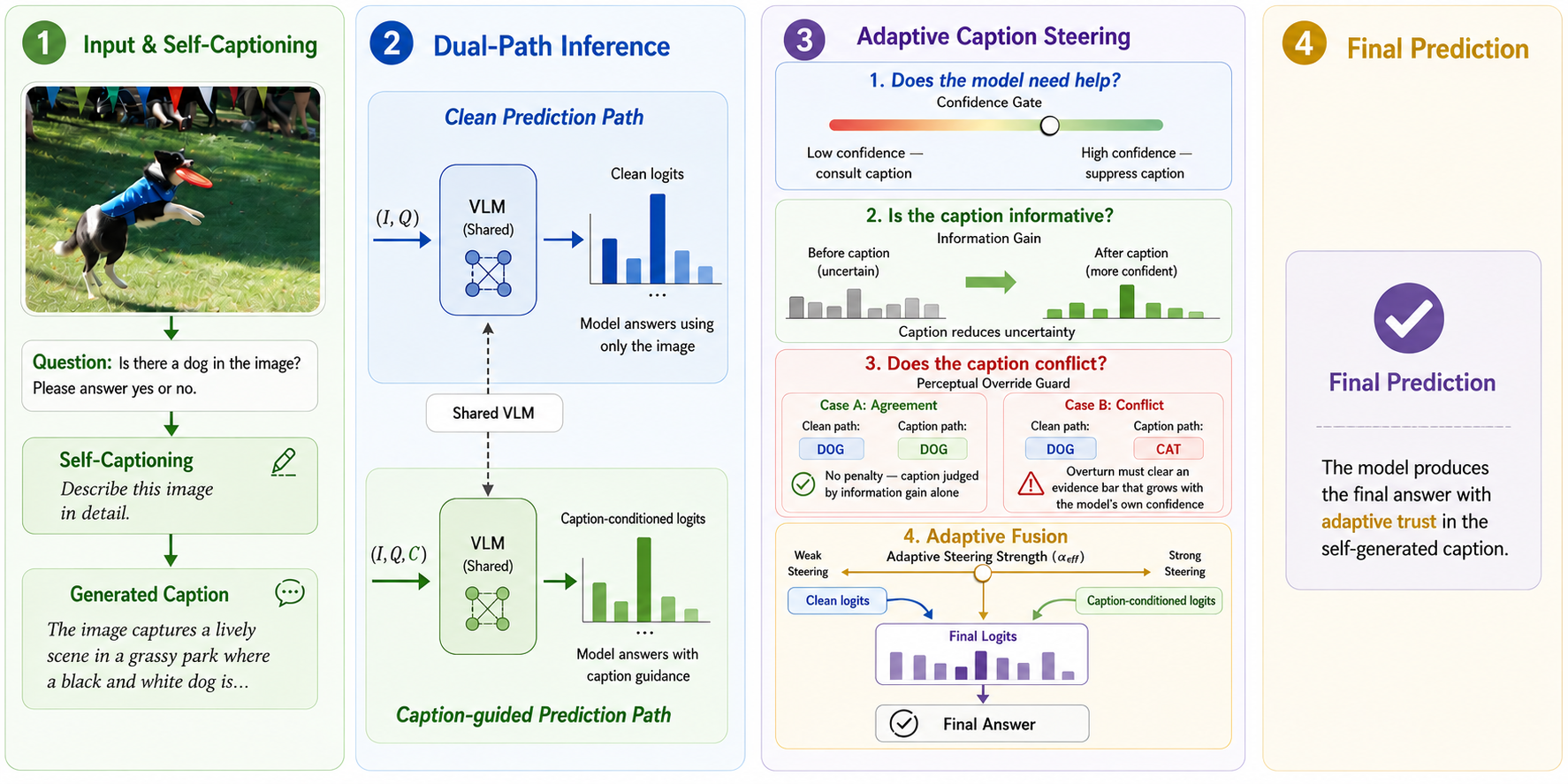}
\caption{Overview of the GEASS pipeline. Given an image $I$ and a question $Q$, the model first generates a caption $C$ via self-captioning (\textbf{Stage 1}). Two parallel forward passes through the same VLM produce logit vectors $\mathbf{z}_{\text{clean}}$ (conditioned on $I,Q$) and $\mathbf{z}_{\text{cap}}$ (conditioned on $I,Q,C$) (\textbf{Stage 2}). The fusion module (\textbf{Stage 3}) computes a confidence gate $\alpha$ (does the model need help?) and a weight $w$ combining an information-gain term (does the caption cover the query?) with a perceptual override guard (how readily may the caption overturn the clean answer?), yielding the effective weight $\alpha_{\text{eff}}=\alpha\cdot w$. The final logit $\mathbf{z}_{\text{final}}$ is decoded into the answer (\textbf{Stage 4}).}
\label{fig:pipeline}
\end{figure*}

Section~\ref{sec:signals} showed that the help/harm regime is exposed by three quantities a decoder already produces at the answer step: the model's confidence on the clean (image-only) path, the entropy reduction a caption induces, and whether the caption overturns the clean answer. GEASS turns these three readings into a per-query control over how much of the caption the model consumes. It runs two forward passes, with and without the caption, and fuses their logits with a query-specific weight assembled from three components: a confidence gate that decides \emph{whether} the caption is consulted, an information-gain term that measures \emph{how much} of the query the caption covers, and a perceptual override guard that controls \emph{how readily} the caption may overturn a visually grounded answer. All three read decoding logits alone---no attention access, no grounding, no extra model---which is what makes the mechanism of Section~\ref{sec:mechanism} actionable at deployment. The pipeline is shown in Figure~\ref{fig:pipeline}.

\subsection{Dual-Path Inference}
\label{sec:dual}

Caption $C$ is produced by the same VLM under the prompt \textit{``Describe this image in detail''} with greedy decoding, fixed once per image. At decoding step $t$, two logit vectors are obtained:
\begin{align}
    \mathbf{z}_{\text{clean}}^{(t)} &= f_\theta(\cdot \mid I, Q, y_{<t}),
    \label{eq:clean}\\
    \mathbf{z}_{\text{cap}}^{(t)}   &= f_\theta(\cdot \mid I, Q, C, y_{<t}),
    \label{eq:cap}
\end{align}
where $y_{<t}$ are the tokens generated so far. The clean path is the model's own answer from the image alone; the caption path is the answer the model would give if it accepted $C$ at face value. The shift $\mathbf{z}_{\text{cap}}^{(t)}-\mathbf{z}_{\text{clean}}^{(t)}$ isolates the caption's contribution, and the rest of this section specifies how much of it to keep.

\subsection{Confidence Gate}
\label{sec:confidence}

A confident clean path means the model has already resolved the query from the image---the regime where, as Section~\ref{sec:phenomenon} showed, a caption can only distract. The gate therefore consults the caption only when the clean path is uncertain. We use the maximum probability of the clean distribution as the confidence proxy,
\begin{equation}
    c^{(t)} = \max_{v \in \mathcal{V}} \, p_{\text{clean}}^{(t)}(v),
    \quad
    p_{\text{clean}}^{(t)} = \mathrm{softmax}\big(\mathbf{z}_{\text{clean}}^{(t)}\big),
    \label{eq:confidence}
\end{equation}
which lies in $(0,1]$ and is comparable across vocabularies and model scales. A sigmoid gate maps it to a fusion coefficient,
\begin{equation}
    \alpha^{(t)} = \sigma\!\big(-\beta(c^{(t)}-\tau)\big),
    \label{eq:alpha}
\end{equation}
with threshold $\tau$ and sharpness $\beta$. The gate is closed ($\alpha\!\to\!0$) when $c^{(t)}\!\gg\!\tau$, where the model is sure of a detail it can read, and open ($\alpha\!\to\!1$) when $c^{(t)}\!\ll\!\tau$, where it needs help; near $\tau$ it interpolates smoothly. We set $\beta$ large enough that the transition occupies a narrow band around $\tau$, so the gate acts as a soft switch rather than a continuous re-weighting.

\subsection{Evidence-Aware Weighting}
\label{sec:evidence}

An open gate exposes the model to the caption but does not decide whether the caption deserves weight. Two questions remain: whether the caption covers the current query, and whether it is reinforcing or overturning the model's own answer. We capture them with two terms that combine into a single weight.

\paragraph{Information gain.}
A caption that covers the query sharpens the prediction; one that omits it leaves the distribution unchanged or more diffuse. We read coverage, without parsing the caption, as the relative entropy reduction
\begin{equation}
    r^{(t)} = \frac{H\big(p_{\text{clean}}^{(t)}\big) - H\big(p_{\text{cap}}^{(t)}\big)}
                   {H\big(p_{\text{clean}}^{(t)}\big) + \epsilon},
    \label{eq:ratio}
\end{equation}
with $H(p)=-\sum_v p(v)\log p(v)$ and a small $\epsilon$ for stability. $r^{(t)}\!>\!0$ means the caption sharpens the model's belief; $r^{(t)}\!\leq\!0$ means it adds noise. Section~\ref{sec:signals} verifies that $r^{(t)}$ is large on queries the caption covers and near zero on those it omits.

\paragraph{Perceptual override guard.}
Information gain is blind to direction: it rewards any sharpening, including a caption that confidently sharpens toward overturning a correct visual answer. Section~\ref{sec:signals} showed that such harmful overturns concentrate where the clean path is confident, so overturning a visually grounded answer should demand evidence in proportion to that confidence. We encode the direction with an agreement indicator,
\begin{equation}
    a^{(t)} = \mathbbm{1}\!\Big[
        \arg\max p_{\text{clean}}^{(t)} = \arg\max p_{\text{cap}}^{(t)}
    \Big],
    \label{eq:agree}
\end{equation}
and scale the override penalty by the clean confidence $c^{(t)}$:
\begin{equation}
    w^{(t)} = \sigma\!\big(\eta\, r^{(t)} - \lambda\,(1-a^{(t)})\,c^{(t)}\big).
    \label{eq:w}
\end{equation}
Under agreement ($a^{(t)}\!=\!1$) the penalty vanishes and $w^{(t)}$ reduces to a sigmoid of scaled information gain: a caption that reinforces the model is admitted on its information alone. Under disagreement ($a^{(t)}\!=\!0$) the caption must clear a confidence-scaled bar---the break-even $w^{(t)}\!=\!\tfrac{1}{2}$ is reached only when $r^{(t)} \geq (\lambda/\eta)\,c^{(t)}$, so the required entropy reduction grows with how firmly the model held its visual answer. With defaults $\eta=5,\lambda=2$, overturning a fully confident answer ($c\!=\!1$) needs a $40\%$ entropy reduction, while a barely-held answer (small $c$) is overturned almost freely---the asymmetry of Figure~\ref{fig:flip_vs_conf}. Confirming the model is cheap; overturning it is expensive, and the price scales with its visual certainty.

\subsection{Logit Fusion}
\label{sec:decoding}

The gate and the weight multiply into the effective fusion weight,
\begin{equation}
    \alpha_{\text{eff}}^{(t)} = \alpha^{(t)} \cdot w^{(t)},
\end{equation}
which is suppressed whenever \emph{either} the model does not need help \emph{or} the caption does not deserve to be trusted. The final logits are a gated residual on top of the clean path,
\begin{equation}
    \mathbf{z}_{\text{final}}^{(t)}
    = \mathbf{z}_{\text{clean}}^{(t)}
      + \alpha_{\text{eff}}^{(t)}
        \big(\mathbf{z}_{\text{cap}}^{(t)} - \mathbf{z}_{\text{clean}}^{(t)}\big),
    \label{eq:fusion}
\end{equation}
and the next token is $y_t=\arg\max_v \mathbf{z}_{\text{final}}^{(t)}(v)$. The two paths keep separate KV caches but share parameters; the selected token is appended to both before the next step. For yes/no and multiple-choice tasks the procedure reduces to a single decoding step on the answer token. GEASS adds one caption-generation pass per image (amortized across its
queries) and one extra forward pass per decoding step for
$\mathbf{z}_{\text{cap}}$ ($\sim$$2.3\times$ greedy decoding overall,
Table~\ref{tab:overhead})
\section{Experiments}
\label{sec:exp}

\subsection{Setup}

\paragraph{Models.} We evaluate GEASS on four VLMs spanning architectures and scales: InternVL2-8B~\cite{chen2024expanding}, InternVL3-8B~\cite{zhu2025internvl3}, Qwen2.5-VL-3B~\cite{bai2025qwen25vl}, and a reasoning-augmented variant fine-tuned with chain-of-thought supervision, denoted Qwen2.5-VL-3B$^\dagger$. All experiments run on a single NVIDIA H200 GPU.

\paragraph{Benchmarks.} We report results on two benchmarks, chosen so that
both the global and the detail regimes of Section~\ref{sec:phenomenon} are
represented. \textbf{POPE}~\cite{li2023evaluating} probes object existence
with yes/no questions under three negative-sampling strategies of increasing
difficulty---\emph{random}, \emph{popular}, and \emph{adversarial}---built
on three source datasets: MSCOCO~\cite{lin2014coco},
A-OKVQA~\cite{schwenk2022okvqa}, and GQA~\cite{hudson2019gqa}. We evaluate
all nine settings and report, per source dataset, the F1 averaged over the
three splits; the full per-split breakdown is in
Appendix~\ref{app:results}. \textbf{HallusionBench}~\cite{guan2024hallusionbench}
targets visual illusions and entangled language--vision hallucinations; we
report all-question accuracy (aAcc), question-pair accuracy (qAcc), and
figure accuracy (fAcc). Both tasks are answered in a single decoding step,
so GEASS is applied once at the answer token.

\paragraph{Baselines.} We compare GEASS against four references: \textbf{Base}, the vanilla model with no intervention; \textbf{Base+Cap}, the same model with the self-generated caption naively appended to the input; \textbf{VCD}~\cite{leng2024mitigating}, which contrasts the output against a noise-corrupted image; and \textbf{CODE}~\cite{kim2024code}, which contrasts against the model's own self-generated description. \textbf{Base+Cap} measures unconditional caption consumption, while VCD and CODE represent the ``subtract the auxiliary signal'' strategy. All methods use greedy decoding. GEASS uses a single fixed setting ($\tau{=}0.5$, $\beta{=}10$, $\eta{=}5$, $\lambda{=}2$) for every model and benchmark; hyperparameter sensitivity, inference overhead, and a comparison against the oracle attention intervention of Section~\ref{sec:mechanism} are deferred to the appendix.

\begin{table*}[t]
\centering
\scriptsize
\setlength{\tabcolsep}{3.5pt}
\renewcommand{\arraystretch}{1.15}
\caption{Main results across four VLMs. For each POPE source dataset (MSCOCO, A-OKVQA, GQA; overall split) we report Accuracy, Recall, and F1 (\%); for HallusionBench we report all-question accuracy (aAcc), question-pair accuracy (qAcc), and figure accuracy (fAcc). $^\dagger$ denotes the reasoning-augmented variant; \emph{Base+Cap} naively appends the self-generated caption. The best entry per (model, column) is \colorbox{gray!25}{\textbf{highlighted}}; -- denotes not evaluated.}
\label{tab:main}
\begin{tabular*}{\textwidth}{@{\extracolsep{\fill}} l l ccc ccc ccc ccc @{}}
\toprule
\multirow{2}{*}{Model} & \multirow{2}{*}{Method}
 & \multicolumn{3}{c}{\textbf{POPE (MSCOCO)}}
 & \multicolumn{3}{c}{\textbf{POPE (A-OKVQA)}}
 & \multicolumn{3}{c}{\textbf{POPE (GQA)}}
 & \multicolumn{3}{c}{\textbf{HallusionBench}} \\
\cmidrule(lr){3-5}\cmidrule(lr){6-8}\cmidrule(lr){9-11}\cmidrule(lr){12-14}
 & & Acc.$\uparrow$ & Rec.$\uparrow$ & F1$\uparrow$
   & Acc.$\uparrow$ & Rec.$\uparrow$ & F1$\uparrow$
   & Acc.$\uparrow$ & Rec.$\uparrow$ & F1$\uparrow$
   & aAcc$\uparrow$ & qAcc$\uparrow$ & fAcc$\uparrow$ \\
\midrule

% ===================== InternVL2-8B =====================
\multirow{5}{*}{InternVL2-8B}
 & Base     & 85.7 & 78.9 & 84.5 & 87.0 & 82.2 & 86.0 & 84.8 & 77.4 & 83.4 & -- & -- & -- \\
 & Base+Cap & 85.1 & 75.2 & 84.1 & 86.4 & 78.6 & 85.3 & 84.1 & 74.3 & 83.3 & -- & -- & -- \\
 & VCD      & 85.8 & \cellcolor{gray!25}\textbf{79.6} & 84.3 & 87.4 & \cellcolor{gray!25}\textbf{82.7} & 86.7 & 85.0 & \cellcolor{gray!25}\textbf{78.1} & 83.7 & -- & -- & -- \\
 & CODE     & 86.0 & 78.0 & 84.7 & 86.7 & 81.1 & 85.9 & 84.3 & 75.9 & 82.9 & -- & -- & -- \\
 & \textbf{GEASS} & \cellcolor{gray!25}\textbf{86.8} & 78.3 & \cellcolor{gray!25}\textbf{85.3} & \cellcolor{gray!25}\textbf{88.1} & 81.9 & \cellcolor{gray!25}\textbf{87.1} & \cellcolor{gray!25}\textbf{85.6} & 78.0 & \cellcolor{gray!25}\textbf{84.8} & -- & -- & -- \\
\midrule

% ===================== InternVL3-8B =====================
\multirow{5}{*}{InternVL3-8B}
 & Base     & 90.4 & 82.8 & 89.9 & 92.0 & 89.0 & 91.6 & 89.9 & 86.2 & 89.6 & 66.71 & 43.50 & 41.84 \\
 & Base+Cap & 90.7 & 80.0 & 90.2 & 91.1 & 85.6 & 90.9 & 89.1 & 82.2 & 89.0 & 59.88 & 34.17 & 33.62 \\
 & VCD      & 91.1 & \cellcolor{gray!25}\textbf{83.7} & \cellcolor{gray!25}\textbf{90.6} & 92.2 & 89.4 & 91.0 & 90.1 & 86.9 & 89.9 & 67.24 & 44.05 & 42.51 \\
 & CODE     & 90.2 & 82.3 & 89.1 & 91.4 & 87.1 & 90.2 & 90.2 & \cellcolor{gray!25}\textbf{87.0} & 90.1 & -- & -- & -- \\
 & \textbf{GEASS} & \cellcolor{gray!25}\textbf{91.4} & 83.5 & 90.5 & \cellcolor{gray!25}\textbf{92.5} & \cellcolor{gray!25}\textbf{89.6} & \cellcolor{gray!25}\textbf{91.8} & \cellcolor{gray!25}\textbf{91.0} & 86.8 & \cellcolor{gray!25}\textbf{90.9} & \cellcolor{gray!25}\textbf{68.87} & \cellcolor{gray!25}\textbf{45.66} & \cellcolor{gray!25}\textbf{43.73} \\
\midrule

% ===================== Qwen2.5-VL-3B =====================
\multirow{5}{*}{Qwen2.5-VL-3B}
 & Base     & 85.8 & 78.0 & 85.2 & 87.3 & 81.4 & 86.6 & 84.8 & \cellcolor{gray!25}\textbf{77.2} & 83.9 & -- & -- & -- \\
 & Base+Cap & 85.0 & 75.5 & 84.8 & 85.9 & 77.9 & 85.4 & 84.8 & 75.4 & 84.2 & -- & -- & -- \\
 & VCD      & 86.5 & \cellcolor{gray!25}\textbf{78.9} & 86.2 & 87.5 & \cellcolor{gray!25}\textbf{81.8} & 86.5 & 84.8 & 76.9 & 84.1 & -- & -- & -- \\
 & CODE     & 86.0 & 77.8 & 85.8 & 87.0 & 80.9 & 86.5 & 84.2 & 75.7 & 83.4 & -- & -- & -- \\
 & \textbf{GEASS} & \cellcolor{gray!25}\textbf{87.0} & 77.4 & \cellcolor{gray!25}\textbf{86.8} & \cellcolor{gray!25}\textbf{88.4} & 81.2 & \cellcolor{gray!25}\textbf{87.9} & \cellcolor{gray!25}\textbf{85.4} & 76.9 & \cellcolor{gray!25}\textbf{84.6} & -- & -- & -- \\
\midrule

% ===================== Qwen2.5-VL-3B (reasoning) =====================
\multirow{5}{*}{Qwen2.5-VL-3B$^\dagger$}
 & Base     & 85.5 & 77.7 & 85.1 & 86.9 & 80.8 & 86.4 & 84.6 & 76.8 & 84.1 & 61.19 & 34.06 & 36.12 \\
 & Base+Cap & 84.8 & 74.8 & 84.2 & 85.8 & 77.0 & 85.5 & 83.8 & 74.6 & 84.0 & 51.31 & 26.48 & 28.83 \\
 & VCD      & 86.2 & \cellcolor{gray!25}\textbf{78.3} & 85.8 & 87.2 & \cellcolor{gray!25}\textbf{81.3} & 86.6 & 84.9 & 77.0 & 84.3 & 63.19 & 35.77 & 37.94 \\
 & CODE     & 85.9 & 77.4 & 85.3 & 87.1 & 80.5 & 86.5 & 85.1 & \cellcolor{gray!25}\textbf{77.3} & \cellcolor{gray!25}\textbf{84.8} & -- & -- & -- \\
 & \textbf{GEASS} & \cellcolor{gray!25}\textbf{87.1} & 78.1 & \cellcolor{gray!25}\textbf{86.6} & \cellcolor{gray!25}\textbf{88.0} & 81.0 & \cellcolor{gray!25}\textbf{87.7} & \cellcolor{gray!25}\textbf{85.8} & 76.5 & 84.7 & \cellcolor{gray!25}\textbf{64.77} & \cellcolor{gray!25}\textbf{37.28} & \cellcolor{gray!25}\textbf{39.12} \\
\bottomrule
\end{tabular*}
\end{table*}

\subsection{Main Results}

Table~\ref{tab:main} reports accuracy across the four benchmarks and four models. Three patterns matter.

\paragraph{Naive caption consumption is broadly harmful.} \emph{Base+Cap} falls well below \emph{Base} on HallusionBench---$66.71\!\to\!59.88$ on InternVL3 and $61.19\!\to\!51.31$ on Qwen2.5-VL-3B$^\dagger$---generalizing the single-number drop of Table~\ref{tab:caption_effect} across models. The harm concentrates where questions hinge on fine content the caption omits, consistent with the mechanism of Section~\ref{sec:mechanism}: the caption pulls attention off the image and the model loses detail it could otherwise read.

\paragraph{GEASS turns the same caption into a net gain.} Reading the identical self-generated caption, GEASS improves over \emph{Base} on every benchmark and recovers the loss that \emph{Base+Cap} incurs. The gain is largest on HallusionBench, where the model's own visual judgment is least reliable and the confidence gate opens most often, and smallest on benchmarks where the clean path is already decisive, where the gate stays closed and GEASS reduces to the baseline by design.

\paragraph{Regulating beats subtracting.} GEASS surpasses VCD and CODE on the benchmarks that mix global and detail questions; the one exception is the InternVL3 POPE split, where VCD's noise contrast edges GEASS on an already-saturated metric. CODE consumes the same self-generated description but subtracts its influence, which discards the global evidence the description does carry; GEASS instead keeps that evidence on the queries it covers and suppresses it only where the caption is uninformative or tries to override a confident answer. The gap is clearest on benchmarks that mix global and detail questions, where a uniform subtract-or-consume policy cannot be right for both.

%% ─────────────────────────────────────────────────────────────────
%% DETAIL/GLOBAL VALIDATION TABLE — single column
%% ─────────────────────────────────────────────────────────────────
\begin{table*}[t]
\centering
\footnotesize
\setlength{\tabcolsep}{10pt}
\renewcommand{\arraystretch}{1.25}
\caption{Component ablation on Qwen2.5-VL-3B$^\dagger$ (\%). POPE F1 is averaged over the three negative-sampling splits of POPE (MSCOCO)
(random, popular, adversarial); HallusionBench is all-question accuracy (aAcc). Removing any component degrades accuracy; removing the information-gain term drops HallusionBench \emph{below} the no-caption \emph{Base} (POPE F1 85.15, aAcc 61.19)---an untargeted caption is worse than none. Best per column in \textbf{bold}.}
\label{tab:ablation}
\vspace{4pt}
\begin{tabular}{l cc}
\toprule
Variant & POPE (F1)\,$\uparrow$ & HallusionBench (aAcc)\,$\uparrow$ \\
\midrule
\rowcolor{gray!10}\textbf{GEASS (full)}                  & \textbf{86.68} & \textbf{64.77} \\
\quad w/o confidence gate                                & 85.43 & 64.13 \\
\quad w/o information gain                                & 86.02 & 60.84 \\
\quad w/o override guard ($\lambda{=}0$)                 & 86.21 & 63.94 \\
\quad override guard w/o confidence scaling ($\times c$) & 86.34 & 62.71 \\
\bottomrule
\end{tabular}
\end{table*}

\subsection{Does GEASS Protect Detail While Keeping Global?}
\label{sec:detail_global}

The main table shows aggregate gains; we now check that they come from the mechanism we claim rather than from a uniform shift. Table~\ref{tab:detail_global} reports accuracy on the global and detail splits of GD-Probe (Section~\ref{sec:phenomenon}). \emph{Base+Cap} reproduces the diagnosis exactly: it raises global accuracy by importing scene-level evidence and lowers detail accuracy by pulling the model off content the caption omits. GEASS retains the global improvement while restoring detail accuracy close to (or above) \emph{Base}. The method therefore does what the analysis predicts---admit the caption where it complements the model and hold it back where it would overwrite perception---rather than trading one regime for the other.

%% ─────────────────────────────────────────────────────────────────
%% ABLATION TABLE — single column
%% ─────────────────────────────────────────────────────────────────
\begin{table}[t]
\centering
\caption{Accuracy (\%) on the global and detail splits (GD-Probe, averaged over the four models). \emph{Base+Cap} helps global questions but hurts detail; GEASS keeps the global gain while recovering detail---the behavior the diagnosis in Section~\ref{sec:phenomenon} predicts.}
\label{tab:detail_global}
\vspace{8pt}
\setlength{\tabcolsep}{10pt}
\renewcommand{\arraystretch}{1.25}
\begin{tabular}{l c c}
\toprule
Method & Global & Detail \\
\midrule
Base       & 65.4 & 70.8 \\
Base+Cap   & 70.1 & 63.2 \\
\rowcolor{gray!12}\textbf{GEASS} & 69.9 & 71.2 \\
\bottomrule
\end{tabular}
\end{table}

\subsection{Ablation}
\label{sec:ablation}

Table~\ref{tab:ablation} removes one component at a time. \textbf{Without the
confidence gate} ($\alpha\!\equiv\!1$) the caption is consulted on every
query, including confident detail questions where it can only distract; the
largest drop is on POPE, whose object-existence queries are precisely those
the clean path usually resolves on its own. \textbf{Without information
gain} (the weight reduces to the override term alone) irrelevant captions
are admitted with equal weight, injecting noise on queries they do not
cover; the effect is sharpest on HallusionBench, which falls below the
no-caption \emph{Base}---an untargeted caption is worse than none.
\textbf{Without the override guard} ($\lambda{=}0$) a caption that
confidently sharpens toward a different answer is no longer held back, so
confident visual answers are overturned and both metrics give back part of
the gain. Finally, \textbf{replacing the confidence-scaled penalty with a
constant one} ($\sigma(\eta r-\lambda(1-a))$ instead of
$\sigma(\eta r-\lambda(1-a)c)$) applies the same override bar to confident
and unconfident answers alike; it under-protects confident answers on POPE
and over-restricts corrections of unconfident ones on HallusionBench, which
loses much of its gain, confirming that scaling the bar by visual confidence
(Section~\ref{sec:evidence}) is what couples the guard to the regime it is
meant to govern. Each component addresses a distinct failure, and the full
method is needed to cover all three.
\section{Conclusion}
\label{sec:conclusion}
We revisited the common assumption that a self-generated caption, once
available, is evidence a VLM should consume, and showed instead that a
caption's usefulness is a per-query property, not a per-corpus one. A single
mechanism explains both directions of the effect: an embedded caption
competes with the image for attention and draws the model's evidence---and
its output tokens---onto its own text, which imports missed evidence on
queries the caption covers and replaces a readable region with silent text
on those it omits, the common case for detail questions. Crucially, this
help/harm regime is exposed by quantities a decoder already produces: the
model's confidence on the image-only path, the entropy reduction the caption
induces, and whether the caption overturns the clean answer.

We turned these three readings into GEASS, a training-free module that
regulates, per query, how much of the caption the model consumes: it gates
the caption by the clean path's confidence, weights its contribution by the
entropy reduction it produces, and---encoding the prior that omissions far
outnumber fabrications---raises the evidence bar for an overturn in
proportion to the model's visual confidence, so confirming the model is
cheap and overturning it expensive. The procedure adds one caption-generation
pass per image and one extra forward pass per query
($\sim$2.3$\times$ greedy decoding, Table~\ref{tab:overhead}), introduces no
trainable parameters, and applies to any VLM that exposes decoding logits.
Across four VLMs and two benchmarks---POPE over nine settings spanning three
source datasets, and HallusionBench---a single fixed hyperparameter setting
improves over both vanilla inference and contrastive decoding, and the gains
come from protecting detail while preserving the global benefit rather than
from a uniform shift.

\paragraph{Limitations and future work.}
GEASS reads agreement between the two pathways as evidence of caption
reliability and penalizes disagreement with a confidence-scaled bar. This is
well calibrated when fabrications are rare relative to omissions---the
regime our analysis documents---but it breaks down when a caption
confidently fabricates: there the override guard would suppress a correction
the model should accept. The method also relies on logit-level proxies for
coverage, which can misfire when the caption sharpens the distribution for
reasons unrelated to genuine evidence. Finally, our evaluation covers
discriminative benchmarks where GEASS reduces to a single decoding step at
the answer token; the per-token formulation of Section~\ref{sec:decoding}
applies unchanged to free-form generation, but validating it on generative
tasks remains future work, as does lifting the rarity-of-fabrication
assumption---for instance through explicit object-level verification that
separates an omission from a fabrication.


\bibliography{example_paper}
\bibliographystyle{icml2025}

%%%%%%%%%%%%%%%%%%%%%%%%%%%%%%%%%%%%%%%%%%%%%%%%%%%%%%%%%%%%%%%%%%%%%%%%%%%%%%%
%%%%%%%%%%%%%%%%%%%%%%%%%%%%%%%%%%%%%%%%%%%%%%%%%%%%%%%%%%%%%%%%%%%%%%%%%%%%%%%
% APPENDIX
%%%%%%%%%%%%%%%%%%%%%%%%%%%%%%%%%%%%%%%%%%%%%%%%%%%%%%%%%%%%%%%%%%%%%%%%%%%%%%%
%%%%%%%%%%%%%%%%%%%%%%%%%%%%%%%%%%%%%%%%%%%%%%%%%%%%%%%%%%%%%%%%%%%%%%%%%%%%%%%
\newpage
\appendix
\onecolumn
\appendix

\section{GD-Probe: Construction and Statistics}
\label{app:gdprobe}
GD-Probe isolates the effect of \emph{question type} from that of the image:
because the global and the detail question share the same picture, and
therefore the same self-generated caption, any difference in caption effect
between them cannot be attributed to the image or to caption quality, only to
what each question asks. This is what licenses the per-query reading of
caption utility in Section~\ref{sec:phenomenon}, and the construction is built
on COCO~\cite{lin2014coco} so that the same object boxes drive the
attention ($\mathrm{TRA}$, \S\ref{sec:mechanism}) and coverage
(\S\ref{sec:coverage}) analyses.

\paragraph{Construction.}
We start from $425$ COCO validation images that carry instance-level box and
label annotations, and form two questions per image---one global and one
detail---for $850$ image--question pairs in total. The \emph{global}
question probes scene-level content---the dominant object, a relation between
dominant objects, the overall layout, or a count---which the annotations
confirm is present and salient. The \emph{detail} question targets a single
annotated object whose ground-truth box covers under $5\%$ of the image area
and lies away from the image center, i.e.\ exactly the kind of small,
peripheral content a saliency-weighted caption is prone to drop
(\S\ref{sec:coverage}). Each question is posed in closed form---yes/no or a
short categorical label evaluable by exact match on the answer token---so that
accuracy is read from a single decoding step, matching the protocol used
throughout Sections~\ref{sec:analysis} and~\ref{sec:exp}. Questions are
generated from the annotations with templates, balanced between positive and
negative answers to remove a yes-bias, and manually checked for answerability
from the image alone. The detail target's box is retained for $\mathrm{TRA}$
and for the covering/omitting label of Table~\ref{tab:coverage_split}.

% FIGURE — 2-3 COCO images, each annotated with its paired global question
% (scene-level) and detail question (small off-center target object), with the
% detail target's GT box drawn.
\begin{figure}[t]
\centering
\includegraphics[width=\linewidth]{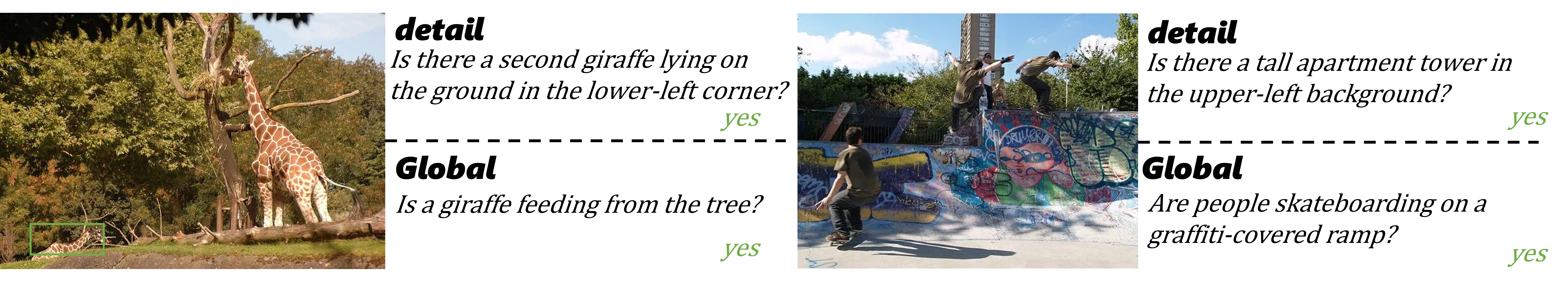}
\caption{Example GD-Probe pairs. \emph{Left:} a giraffe feeding from a tree
(global) paired with a second giraffe lying in the lower-left corner (detail).
\emph{Right:} people skateboarding on a graffiti-covered ramp (global) paired
with a tall tower in the upper-left background (detail). The detail target's
ground-truth box is drawn. The image and its caption are held fixed across the
two questions, so the contrast measures the question, not the picture.}
\label{fig:gdprobe_examples}
\end{figure}
\begin{table}[t]
\centering
\caption{GD-Probe statistics. Each image contributes one global and one
detail question; detail targets are small ($<5\%$ of image area) and
off-center by construction.}
\label{tab:gdprobe_stats}
\vspace{8pt}
\footnotesize
\setlength{\tabcolsep}{6pt}
\renewcommand{\arraystretch}{1.15}
\begin{tabular}{l c}
\toprule
Property & Value \\
\midrule
Source images (COCO val)                 & $425$ \\
Image--question pairs                     & $850$ \\
\quad Global questions                    & $425$ \\
\quad Detail questions                    & $425$ \\
Mean detail-target box area (\% of image) & $2.1$ \\
Max detail-target box area (\% of image)  & $<5$ \\
Answer balance (\% positive)              & $50.0$ \\
\bottomrule
\end{tabular}
\end{table}
\section{Attention Analysis: Definitions, Robustness, and the Oracle Intervention}
\label{app:attention}
This section makes the attention measurements of \S\ref{sec:mechanism} precise,
shows the conclusions are insensitive to the aggregation choice, and specifies
the oracle $\gamma$-intervention whose comparison against GEASS is reported in
Appendix~\ref{app:results}.

\paragraph{Aggregation and metrics.}
All attention quantities are read at the answer step. For a chosen set of
decoder layers we average the answer-token attention over heads within each
layer and then over layers, giving a single weight $a_j$ per context token,
which we partition into image tokens $\mathcal{I}$, caption tokens
$\mathcal{C}$, and the remainder. $\mathrm{IAS}$, $\mathrm{CAS}$, and
$\mathrm{TRA}$ are then as defined in \S\ref{sec:mechanism}; the default reads
the upper third of the decoder layers, where answer-token attention is most
concentrated on content tokens. The Output--Caption Overlap of
Figure~\ref{fig:oco_cas} is the fraction of response $n$-grams
($n\in\{1,2,3\}$) that also occur in the caption, averaged over the response.
Because the diagnostic requires the attention weights themselves, we extract
them with an eager (weight-exposing) attention implementation; the deployed
method (Section~\ref{sec:method}) needs none of this.

\paragraph{Layer and head robustness.}
Table~\ref{tab:attn_robust} recomputes the reallocation of
Table~\ref{tab:attention_shift} under different layer bands. The pattern is
qualitatively stable: adding the caption lowers $\mathrm{IAS}$ and raises
$\mathrm{CAS}$, and the target region cools ($\Delta\mathrm{TRA}<0$) far more
for detail than for global questions, in every band. Replacing the per-head
mean with a per-head max leaves the sign and ordering unchanged (not shown).
The effect is therefore a property of the model's answer-token attention, not
of the particular layers or heads we read.

\begin{table}[t]
\centering
\caption{Robustness of the attention reallocation to the layer band
(answer-token attention; head-averaged). The signs and the
detail$>$global ordering of $\Delta\mathrm{TRA}$ hold across bands; the upper
third (default) is reported in Table~\ref{tab:attention_shift}.}
\label{tab:attn_robust}
\vspace{8pt}
\scriptsize
\setlength{\tabcolsep}{4pt}
\begin{tabular}{ll ccc cc}
\toprule
\multirow{2}{*}{Model} & \multirow{2}{*}{Layer band}
 & $\mathrm{IAS}_{\text{clean}}$ & $\mathrm{IAS}_{\text{cap}}$ & $\mathrm{CAS}_{\text{cap}}$
 & $\Delta\mathrm{TRA}$ & $\Delta\mathrm{TRA}$ \\
 & & (\%) & (\%) & (\%) & global & detail \\
\midrule
\multirow{4}{*}{InternVL2-8B}
 & Lower third            & 55.1 & 45.9 & 14.6 & $-1.9$ & $-7.4$ \\
 & Middle third           & 59.7 & 46.8 & 18.6 & $-2.9$ & $-11.1$ \\
 & Upper third (default)  & 62.4 & 47.1 & 22.3 & $-3.8$ & $-14.6$ \\
 & All layers             & 59.0 & 46.5 & 18.4 & $-2.8$ & $-10.8$ \\
\midrule
\multirow{4}{*}{Qwen2.5-VL-3B}
 & Lower third            & 51.3 & 42.4 & 17.1 & $-2.3$ & $-8.7$ \\
 & Middle third           & 55.5 & 43.0 & 21.7 & $-3.5$ & $-12.9$ \\
 & Upper third (default)  & 58.0 & 43.5 & 26.1 & $-4.5$ & $-16.9$ \\
 & All layers             & 54.8 & 43.0 & 21.5 & $-3.4$ & $-12.7$ \\
\bottomrule
\end{tabular}
\end{table}

\paragraph{The oracle $\gamma$-intervention.}
The causal test of \S\ref{sec:mechanism} edits attention directly. At the
answer step we down-weight the caption tokens and renormalize,
\begin{equation}
\tilde a_j =
\begin{cases}
a_j/\gamma, & j\in\mathcal{C},\\
a_j, & \text{otherwise},
\end{cases}
\qquad
\hat a_j = \frac{\tilde a_j}{\sum_k \tilde a_k},
\label{eq:gamma}
\end{equation}
with $\gamma\geq 1$; $\gamma{=}1$ is the unmodified caption condition and
larger $\gamma$ restores image attention. We treat this as an
\emph{oracle}: it needs read/write access to the model's internal attention
(unavailable under the efficient kernels common at deployment) and, in
Appendix~\ref{app:results}, is granted its best operating point as an upper
reference on what attention-level control can recover. GEASS pulls the same lever from the output side, through logit fusion that
requires no access to attention weights, and Appendix~\ref{app:results}
shows it approaches the oracle without any attention access.

\section{Hyperparameter Sensitivity and Inference Overhead}
\label{app:hp}

\paragraph{Sensitivity.}
GEASS has four hyperparameters: the gate threshold $\tau$ and sharpness $\beta$
in the confidence gate $\alpha=\sigma(-\beta(c-\tau))$, the information-gain
scale $\eta$, and the override strength $\lambda$ in the evidence weight
$w=\sigma(\eta r-\lambda(1-a)c)$. Table~\ref{tab:hp_sweep} varies one
hyperparameter at a time around its default with the other three fixed, on
Qwen2.5-VL-3B$^\dagger$. Accuracy is flat over a wide band around the defaults
$(\tau,\beta,\eta,\lambda)=(0.5,10,5,2)$, which is why a single setting transfers
across all four models and benchmarks (Section~\ref{sec:exp}) without per-model
tuning. The trends follow the design: lowering $\tau$ closes the gate on more
queries (fewer captions consulted), raising $\beta$ sharpens the gate into a
switch, and the override behaves through the ratio $\lambda/\eta$, which fixes
the break-even entropy reduction $r=(\lambda/\eta)\,c$ for overturning a clean
answer of confidence $c$ ($0.4$ at the defaults, \S\ref{sec:evidence}). Extreme
values recover known degenerate cases---$\beta{\to}0$ removes the gate,
$\lambda{=}0$ removes the override guard---both already covered by the ablation
in Table~\ref{tab:ablation}. This is a one-at-a-time study around the operating
point, not a full grid, so it measures local sensitivity rather than joint
interactions; the only pair coupled by construction is $(\eta,\lambda)$, through
the ratio $\lambda/\eta$ above, so their two sweeps probe the same mechanism from
opposite sides. A full factorial is omitted for cost, and the per-axis stability
is what justifies the single fixed setting.

\begin{table}[t]
\centering
\footnotesize
\setlength{\tabcolsep}{6pt}
\renewcommand{\arraystretch}{1.15}
\caption{Hyperparameter sensitivity on Qwen2.5-VL-3B$^\dagger$ (\%). POPE F1 is averaged over the three negative-sampling splits of POPE (MSCOCO); HallusionBench is aAcc. Defaults ($\tau{=}0.5,\,\beta{=}10,\,\eta{=}5,\,\lambda{=}2$, in \textbf{bold}) maximize both metrics; -- not yet run.}
\label{tab:hp_sweep}
\vspace{8pt}
\begin{tabular}{ll cc}
\toprule
Param & Value & POPE (F1)\,$\uparrow$ & HallusionBench (aAcc)\,$\uparrow$ \\
\midrule
\multirow{5}{*}{$\tau$}
 & $0.3$          & 86.04 & 63.11 \\
 & $0.4$          & 86.41 & --    \\
 & $\mathbf{0.5}$ & \textbf{86.68} & \textbf{64.77} \\
 & $0.6$          & 86.39 & --    \\
 & $0.7$          & --    & --    \\
\midrule
\multirow{4}{*}{$\beta$}
 & $5$            & --    & --    \\
 & $\mathbf{10}$  & \textbf{86.68} & \textbf{64.77} \\
 & $20$           & 86.62 & --    \\
 & $50$           & 86.55 & --    \\
\midrule
\multirow{3}{*}{$\eta$}
 & $2$            & 86.40 & 63.71 \\
 & $\mathbf{5}$   & \textbf{86.68} & \textbf{64.77} \\
 & $10$           & --    & --    \\
\midrule
\multirow{3}{*}{$\lambda$}
 & $1$            & 86.45 & --    \\
 & $\mathbf{2}$   & \textbf{86.68} & \textbf{64.77} \\
 & $4$            & --    & --    \\
\bottomrule
\end{tabular}
\end{table}

\paragraph{Why logits, not attention.} Our analysis in Section~\ref{sec:mechanism} traces caption-induced errors to a shift in attention away from the image. A natural response is to edit the attention directly, as a large family of methods does (OPERA~\cite{huang2024opera}, PAI~\cite{liu2024paying}, AttnReal~\cite{tu2026attention}). We deliberately avoid this. All such methods must read the model's internal attention weights, which forces the use of \emph{eager} attention instead of fused kernels such as FlashAttention; because modern VLMs feed thousands of visual tokens into the decoder, materializing the attention map is quadratic in sequence length and is precisely the cost that efficient deployment is built to avoid. The strongest attention method, OPERA, compounds this with beam search and retrospection-rollback. GEASS instead operates entirely on the output logits: it requires no attention access, remains compatible with fused kernels, and needs no beam search. Its overhead is one amortizable caption pass plus a single extra forward (Table~\ref{tab:overhead}), and---unlike the attention family---it carries this cost without disabling the kernels that make VLM inference efficient in the first place.

\begin{table}[t]
\centering
\caption{GEASS intervenes at the output logits, not the attention. Every attention-intervention method requires access to internal attention weights, which forces \emph{eager} attention in place of fused kernels (e.g.\ FlashAttention) and scales poorly with the thousands of visual tokens in modern VLMs; OPERA additionally relies on beam search with retrospection-rollback. GEASS reads only logits, keeps fused kernels, and avoids beam search. ``Overhead'' is per-query compute relative to greedy \emph{Base} (Qwen2.5-VL-3B); OPERA is beam-width dependent.}
\label{tab:overhead}
\vspace{8pt}
\footnotesize
\setlength{\tabcolsep}{5pt}
\renewcommand{\arraystretch}{1.25}
\begin{tabular}{l c c c c c}
\toprule
Method & Attn.\ access & Beam search & Fused-kernel & Logit-only & Overhead \\
\midrule
Base (greedy)                  & no  & no  & \textbf{yes} & ---          & $1.0\times$ \\
OPERA~\cite{huang2024opera}    & yes & yes & no           & no           & high \\
PAI~\cite{liu2024paying}          & yes & no  & no           & no           & $\sim$$2\times$ \\
AttnReal~\cite{tu2026attention}& yes & no  & no           & no           & $\sim$$1.1\times$ \\
\rowcolor{gray!12}\textbf{GEASS} & \textbf{no} & no & \textbf{yes} & \textbf{yes} & $\sim$$2.3\times$ \\
\bottomrule
\end{tabular}
\end{table}

\section{Additional Results}
\label{app:results}

\paragraph{Per-cell counts for Table~\ref{tab:coverage_split}.}
Table~\ref{tab:coverage_counts} gives the sample count behind each
type$\times$coverage cell of Table~\ref{tab:coverage_split}. Counts are
per model because coverage is defined on each model's own self-generated
caption, so the covering/omitting partition shifts with the captioner. The
distribution confirms the premise of \S\ref{sec:coverage}: global questions
fall mostly in the covered cells and detail questions mostly in the omitted
ones.

\begin{table}[t]
\centering
\caption{Per-cell sample counts for Table~\ref{tab:coverage_split} (out of 425
global and 425 detail questions per model). Coverage is labeled on each model's
own caption, so counts differ across models. Global queries concentrate in
Covered; detail queries in Omitted.}
\label{tab:coverage_counts}
\vspace{8pt}
\scriptsize
\setlength{\tabcolsep}{5pt}
\begin{tabular}{ll|cccc}
\toprule
\multirow{2}{*}{Type} & \multirow{2}{*}{Coverage}
 & InternVL2 & InternVL3 & Qwen2.5 & Qwen2.5$^\dagger$ \\
\midrule
\multirow{2}{*}{Global} & Covered & 362 & 371 & 353 & 359 \\
                        & Omitted & 63  & 54  & 72  & 66  \\
\midrule
\multirow{2}{*}{Detail} & Covered & 102 & 104 & 92  & 105 \\
                        & Omitted & 323 & 321 & 333 & 320 \\
\bottomrule
\end{tabular}
\end{table}

\paragraph{GEASS vs.\ the oracle intervention.}
Table~\ref{tab:oracle} places GEASS against the oracle $\gamma$-intervention of
Appendix~\ref{app:attention} on the global/detail splits of GD-Probe. The
oracle, which edits internal attention and is given its best operating point,
upper-bounds what attention-level control recovers on detail while leaving
global intact. GEASS attains most of that detail recovery from decoding signals
alone---no boxes, no attention internals---supporting the claim of
\S\ref{sec:mechanism} that the caption's influence is a lever it can pull at the
logit level.

\begin{table}[t]
\centering
\caption{GEASS against the oracle attention intervention
(Appendix~\ref{app:attention}) on GD-Probe (accuracy, \%, averaged over the four
models). The oracle requires attention access and is an upper reference; GEASS
approaches its detail recovery using decoding signals only.}
\label{tab:oracle}
\vspace{8pt}
\footnotesize
\setlength{\tabcolsep}{10pt}
\renewcommand{\arraystretch}{1.15}
\begin{tabular}{l cc}
\toprule
Method & Global & Detail \\
\midrule
Base                          & 65.4 & 70.8 \\
Base+Cap                      & 70.1 & 63.2 \\
\textbf{GEASS}                & 69.9 & 71.2 \\
Oracle $\gamma$-intervention  & 70.3 & 72.5 \\
\bottomrule
\end{tabular}
\end{table}

\section{Qualitative Examples}
\label{app:qual}
\begin{figure}[t]
\centering
\includegraphics[width=0.9\linewidth]{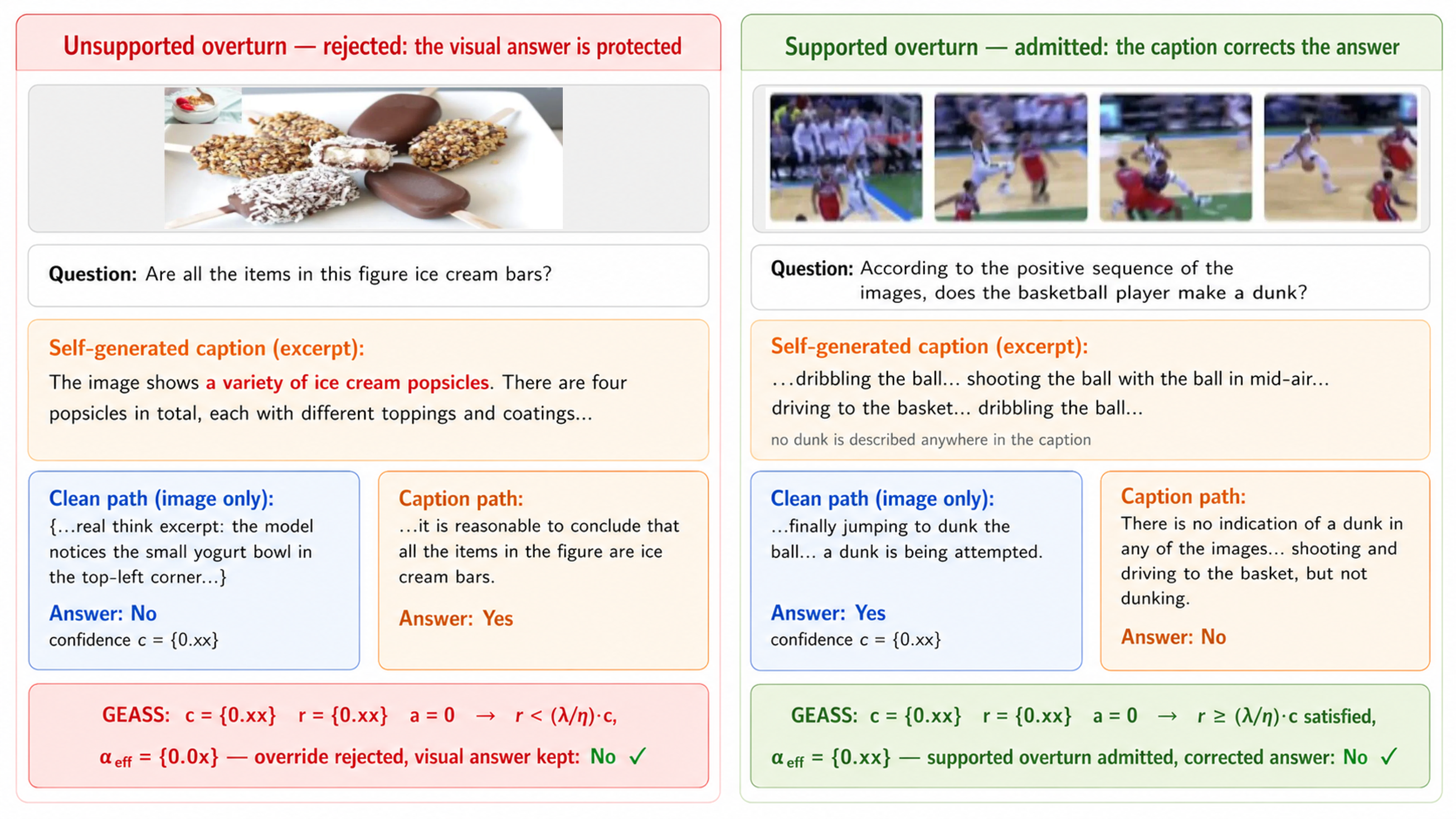}
\caption{Two conflict cases ($a{=}0$) resolved oppositely by the override
guard. \emph{Left:} the clean path is confident and the caption induces
little entropy reduction, so $r<(\lambda/\eta)\,c$: the unsupported
overturn is rejected and the correct visual answer is protected.
\emph{Right:} the clean path is uncertain and the caption sharpens the
prediction enough to clear the confidence-scaled bar
$r\geq(\lambda/\eta)\,c$: the supported overturn is admitted and the
answer is corrected. Signal values match Table~\ref{tab:case_signals}.}
\label{fig:cases}
\end{figure}

Figure~\ref{fig:cases} shows GEASS on representative queries, and
Table~\ref{tab:case_signals} reads off the decoding signals and the resulting
fusion weight for each, tying the behavior back to the three components of
Section~\ref{sec:method}. The cases span the regimes diagnosed in
\S\ref{sec:signals}: a detail query the caption omits, where the clean path is
confident ($c$ high, $r{\approx}0$) so the gate stays nearly closed and the
correct visual answer is protected; a global query the model is unsure of
($c$ low) and the caption covers ($r$ high), where the gate opens and the
caption corrects the answer; and a disagreement on a confident clean answer,
where the override guard demands $r\geq(\lambda/\eta)c$ and rejects an
unsupported overturn while still admitting a well-supported one.

% FIGURE — 3-4 cases. For each: image, question, caption excerpt, the clean
% answer, the caption-conditioned answer, and the GEASS answer (left: caption
% would cause a hallucination, suppressed; right: caption corrects, admitted).
\begin{table}[t]
\centering
\caption{Decoding signals and the resulting effective weight
$\alpha_{\text{eff}}=\alpha\cdot w$ on the cases of
Figure~\ref{fig:cases}. The action column follows from the signals by the
rules of Section~\ref{sec:method}.}
\label{tab:case_signals}
\vspace{8pt}
\scriptsize
\setlength{\tabcolsep}{4pt}
\renewcommand{\arraystretch}{1.2}
\begin{tabular}{l l l ccc c l}
\toprule
Case & Type & Coverage & $c$ & $r$ & $a$ & $\alpha_{\text{eff}}$ & GEASS action \\
\midrule
1 & Detail & Omitted & 0.85 & 0.02 & $0$ & 0.00 & Suppress caption (protect clean) \\
2 & Global & Covered & 0.42 & 0.45 & $1$ & 0.62 & Adopt caption (reinforce) \\
3 & Detail & Omitted & 0.62 & 0.15 & $0$ & 0.09 & Reject unsupported overturn \\
4 & Global & Covered & 0.38 & 0.40 & $0$ & 0.60 & Admit supported correction \\
\bottomrule
\end{tabular}
\end{table}

% \section{Example Cases.}

% \begin{figure}[htbp]
%   \centering
%   \includegraphics[width=0.9\textwidth]{image/case.png}
%   \caption{Left: example where caption causes hallucination (need to reduce caption influence). Right: example where caption corrects the prediction (need normal caption steering).}
%   \label{fig:case_example}
% \end{figure}
%%%%%%%%%%%%%%%%%%%%%%%%%%%%%%%%%%%%%%%%%%%%%%%%%%%%%%%%%%%%%%%%%%%%%%%%%%%%%%%
%%%%%%%%%%%%%%%%%%%%%%%%%%%%%%%%%%%%%%%%%%%%%%%%%%%%%%%%%%%%%%%%%%%%%%%%%%%%%%%

\end{document}